\documentclass{article} 
\usepackage{iclr2027_conference,times}

\usepackage{amsmath,amsfonts,bm}

\def\eqref#1{equation~\ref{#1}}

\def\1{\bm{1}}

\DeclareMathAlphabet{\mathsfit}{\encodingdefault}{\sfdefault}{m}{sl}
\SetMathAlphabet{\mathsfit}{bold}{\encodingdefault}{\sfdefault}{bx}{n}

\usepackage{xcolor}
\usepackage{colortbl}
\usepackage{amsmath}
\usepackage{amsfonts}
\usepackage{amssymb}
\usepackage{bm}
\usepackage{graphicx}
\usepackage{subcaption}
\usepackage{booktabs}
\usepackage{multirow}
\usepackage{arydshln}
\usepackage{wrapfig}
\usepackage{float}
\usepackage{nicefrac}
\usepackage{microtype}
\usepackage{url}
\usepackage{hyperref}
\usepackage[accsupp]{axessibility}  

\title{VHA: VQA-based Hallucination Assessment for Generative Image Super-Resolution}

\author{Donghun Ryou$^{2}$ \qquad   Inju Ha$^{1}$ \qquad Jaeyoo Park$^{1}$ \qquad Bohyung Han$^{1,2}$ \\
Computer Vision Laboratory, $^1$ECE \& $^2$IPAI, Seoul National University\\
{\tt\small \{dhryou, hij1112, bellos1203, bhhan\}@snu.ac.kr}
}

\newcommand{\jy}{\textcolor{magenta}}

\iclrfinalcopy 
\begin{document}

\maketitle
\fancyhead{}
\renewcommand{\headrulewidth}{0pt}

\vspace{-1.5mm}

\begin{abstract}

{Generative image super-resolution produces visually realistic high-resolution images, but may also introduce hallucinations by adding unsupported content or altering existing details.
Because existing super-resolution metrics fail to measure whether generated details remain grounded in the reference image, we present a hallucination assessment technique based on bidirectional visual question answering (VQA), which we term VQA-based Hallucination Assessment (VHA).
The proposed approach uses a vision-language model (VLM) to determine whether the semantic content and low-level visual evidence in a super-resolved image are supported by its ground-truth counterpart.
To this end, it constructs question--answer pairs from each image and evaluates them on both images.
Discrepancies between the resulting scores in both directions reveal missing or altered ground-truth information and unsupported additions.
However, obtaining these scores requires computationally expensive VLM inference.
We therefore develop VHA+, a learned, lightweight model that directly predicts VHA scores from ground-truth and super-resolved image pairs.
We construct a hallucination benchmark, VHA-bench, through a human study and show that both VHA and VHA+ correlate highly with human judgments of hallucination.
The code, benchmark, and model weights are available at \url{https://github.com/dhryougit/Hallucination-Assessment}.}

\end{abstract}

\vspace{-1.5mm}


\section{Introduction}
\label{Introduction}

Generative image super-resolution (SR) can synthesize visually realistic details from low-resolution inputs by drawing on learned image priors~\citep{wang2018esrgan,wang2024exploiting,wei2025perceive,wu2026vosr}.
Since a low-resolution input can correspond to multiple high-resolution images, a plausible reconstruction may differ from the ground truth in its semantic content or visual attributes.
An SR output may alter an object's identity, remove meaningful details, or introduce a texture absent from the original image~\citep{ren2025hallucination,li2025structsr}.
Such discrepancies illustrate hallucinations in generative SR, which we define as the loss or alteration of ground-truth evidence in the super-resolved image, or the addition of content unsupported by the ground truth.

\begin{figure}[h]
    \centering
    \begingroup
    \centering
    \begin{minipage}[t]{0.485\linewidth}
        \centering
        \begin{minipage}[t]{0.32\linewidth}
            \centering
            \small GT\par\vspace{3pt}
            \includegraphics[width=\linewidth]{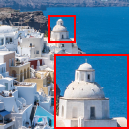}
        \end{minipage}\hfill%
        \begin{minipage}[t]{0.32\linewidth}
            \centering
            \small StableSR\par\vspace{3pt}
            \includegraphics[width=\linewidth]{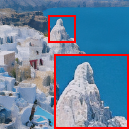}
        \end{minipage}\hfill%
        \begin{minipage}[t]{0.32\linewidth}
            \centering
            \small DiffBIR\par\vspace{3pt}
            \includegraphics[width=\linewidth]{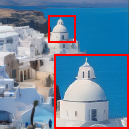}
        \end{minipage}
        \par\vspace{3pt}
        \fontsize{8.5}{9}\selectfont
        \noindent\makebox[0.32\linewidth][c]{PSNR $\uparrow$}\hfill%
        \makebox[0.32\linewidth][c]{\textbf{17.63}}\hfill%
        \makebox[0.32\linewidth][c]{17.11}\par\vspace{0.5pt}
        \noindent\makebox[0.32\linewidth][c]{MUSIQ $\uparrow$}\hfill%
        \makebox[0.32\linewidth][c]{74.14}\hfill%
        \makebox[0.32\linewidth][c]{\textbf{75.18}}\par\vspace{0.5pt}
        \noindent\makebox[0.32\linewidth][c]{VHA (\textbf{Ours}) $\downarrow$}\hfill%
        \makebox[0.32\linewidth][c]{0.6382}\hfill%
        \makebox[0.32\linewidth][c]{\textbf{0.4954}}\par\vspace{0.5pt}
        \noindent\makebox[0.32\linewidth][c]{VHA+ (\textbf{Ours}) $\downarrow$}\hfill%
        \makebox[0.32\linewidth][c]{0.8641}\hfill%
        \makebox[0.32\linewidth][c]{\textbf{0.3488}}\par\vspace{0.5pt}
        \noindent\makebox[0.32\linewidth][c]{Human eval.}\hfill%
        \makebox[0.32\linewidth][c]{\textcolor{gray}{--}}\hfill%
        \makebox[0.32\linewidth][c]{\textcolor{green!45!black}{\raisebox{-0.1ex}[0pt][0pt]{\large$\checkmark$}}}\par\vspace{0.5pt}
    \end{minipage}\hfill%
    \begin{minipage}[t]{0.485\linewidth}
        \centering
        \begin{minipage}[t]{0.32\linewidth}
            \centering
            \small GT\par\vspace{3pt}
            \includegraphics[width=\linewidth]{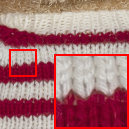}
        \end{minipage}\hfill%
        \begin{minipage}[t]{0.32\linewidth}
            \centering
            \small DiffBIR\par\vspace{3pt}
            \includegraphics[width=\linewidth]{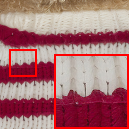}
        \end{minipage}\hfill%
        \begin{minipage}[t]{0.32\linewidth}
            \centering
            \small PURE\par\vspace{3pt}
            \includegraphics[width=\linewidth]{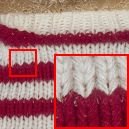}
        \end{minipage}
        \par\vspace{3pt}
        \fontsize{8.5}{9}\selectfont
        \noindent\makebox[0.32\linewidth][c]{PSNR $\uparrow$}\hfill%
        \makebox[0.32\linewidth][c]{\textbf{23.24}}\hfill%
        \makebox[0.32\linewidth][c]{23.14}\par\vspace{0.5pt}
        \noindent\makebox[0.32\linewidth][c]{MUSIQ $\uparrow$}\hfill%
        \makebox[0.32\linewidth][c]{\textbf{70.82}}\hfill%
        \makebox[0.32\linewidth][c]{67.51}\par\vspace{0.5pt}
        \noindent\makebox[0.32\linewidth][c]{VHA (\textbf{Ours}) $\downarrow$}\hfill%
        \makebox[0.32\linewidth][c]{0.5872}\hfill%
        \makebox[0.32\linewidth][c]{\textbf{0.5126}}\par\vspace{0.5pt}
        \noindent\makebox[0.32\linewidth][c]{VHA+ (\textbf{Ours}) $\downarrow$}\hfill%
        \makebox[0.32\linewidth][c]{1.5106}\hfill%
        \makebox[0.32\linewidth][c]{\textbf{0.4791}}\par\vspace{0.5pt}
        \noindent\makebox[0.32\linewidth][c]{Human eval.}\hfill%
        \makebox[0.32\linewidth][c]{\textcolor{gray}{--}}\hfill%
        \makebox[0.32\linewidth][c]{\textcolor{green!45!black}{\raisebox{-0.1ex}[0pt][0pt]{\large$\checkmark$}}}\par\vspace{0.5pt}
    \end{minipage}\par
    \endgroup
    \caption{
        Limitation of conventional evaluation metrics.
        In the left example, PSNR favors a reconstruction that incorrectly depicts the building as a rock formation.
        In the right example, even the perceptual quality metric, MUSIQ~\citep{ke2021musiq} favors a reconstruction that alters the knitted texture present in the ground truth.
        In contrast, VHA and VHA+ assess the hallucinations, which conventional metrics fail to capture.
        }
    \label{fig:teaser_preview}
\end{figure}

Existing SR evaluation metrics do not adequately reflect whether reconstructed content is faithful to the ground truth.
Figure~\ref{fig:teaser_preview} illustrates these limitations with visual examples in which conventional metrics favor less faithful reconstructions.
Distortion metrics such as PSNR and perceptual similarity metrics such as LPIPS~\citep{zhang2018unreasonable} quantify discrepancies in pixel values or learned visual features, but these differences do not necessarily reflect the significance of a content change.
Perceptual quality metrics~\citep{ke2021musiq, wang2023exploring, you2025teaching} can assign favorable scores to visually realistic outputs even when their content is unsupported by the ground truth.

We introduce VQA-based Hallucination Assessment (VHA), which compares how strongly the ground truth and the SR output support individual visual evidence.
Prior work~\citep{ren2025hallucination} produces a hallucination rating and its explanation, but does not explicitly quantify how individual content or attribute changes contribute to the rating.
We represent these evidences as question–answer pairs about specific image content and attributes. 
If the image content is preserved, the same answer should remain supported under both images.
Inspired by VQAScore~\citep{lin2024evaluating}, we measure the changes by comparing the VLM’s answer log-probabilities across the two images on same question-answer pairs.
We aggregate these probability differences into the VHA score.

VHA assesses hallucinations using question–answer pairs generated from both the ground-truth and SR output.
Questions from the ground truth assess whether original content is preserved in the SR output, while questions from the SR output assess whether its content is supported by the ground truth.
In both directions, taxonomy-guided questions cover high-level semantics, such as objects and text, and low-level visual attributes, such as color and texture. 
To reduce the cost of repeated VLM inference, we introduce VHA+, a lightweight model trained on image pairs labeled with VHA scores.
VHA+ predicts these scores directly from ground-truth and SR image pairs without VLM inference.

We evaluate the alignment of VHA and VHA+ with human hallucination judgments. 
We construct VHA-bench, a human-annotated benchmark covering outputs from multiple generative SR methods. 
On this benchmark, VHA and VHA+ achieve higher correlations with human judgments than conventional SR metrics and direct VLM ratings using the same backbones. 
We also show that fine-tuning an SR model on VHA+-based supervision with an image-quality objective both reduces the VHA score and improves image quality metrics.
Our main contributions are as follows:

\begin{itemize}
\item We introduce a bidirectional VQA-based hallucination assessment method (VHA) that compares answer log-probabilities for same question–answer pairs across ground-truth and SR outputs. Taxonomy-guided question–answer generation enables VHA to assess hallucinations across diverse semantic and visual aspects.  
\item We propose VHA+, a lightweight model trained to predict VHA scores without VLM inference, and demonstrate that its learned representations can guide the fine-tuning of a super-resolution model.
\item We construct a human-annotated VHA-bench and demonstrate that VHA and VHA+ correlate stronger with human judgments than conventional SR metrics and direct VLM ratings.
\end{itemize}

\section{Related Works}
\label{sec:relatedworks}

\paragraph{Evaluation Metrics for Super-Resolution.}
SR evaluation commonly considers distortion, perceptual similarity, and perceptual quality.
Distortion metrics such as PSNR and SSIM~\citep{wang2004image} summarize pixel-wise reconstruction errors and local structural differences against a reference, respectively.
LPIPS~\citep{zhang2018unreasonable} and DISTS~\citep{ding2020image} compare deep image representations, with DISTS explicitly combining structure and texture similarity.
No-reference image quality assessment methods, including MUSIQ~\citep{ke2021musiq}, CLIP-IQA~\citep{wang2023exploring}, and DeQA-Score~\citep{you2025teaching}, estimate perceptual quality from the test image alone.
A-FINE~\citep{chen2025toward} combines reference-based comparisons with image naturalness estimates, extending quality prediction to settings where the reference itself may be imperfect.

\paragraph{Hallucination in Generative Image Super-Resolution.}
Generative SR models can produce visually realistic details while introducing semantic errors and spurious structures~\citep{wu2024seesr,wu2026vosr}.
PURE~\citep{wei2025perceive} addresses this issue by integrating image understanding with autoregressive restoration to preserve semantic content and local structures.
However, its evaluation does not separately quantify hallucinations. 
To directly assess hallucinations, \citet{ren2025hallucination} introduce Hallucination Score, which prompts GPT-4o with low-resolution, super-resolved, and ground-truth images to produce an overall hallucination rating and an accompanying explanation.
Although the explanation can identify specific discrepancies, their individual contributions to the overall rating are not explicitly quantified.
VHA constructs its assessment signal by aggregating measured changes in support for individual visual evidences. 

\paragraph{VQA-based Evaluation of Generated Images.}
Visual question answering has been used to assess the semantic alignment and factual correctness of generated images.
TIFA~\citep{hu2023tifa} and VQ$^2$~\citep{yarom2023what} generate questions from textual descriptions and use VQA models to evaluate whether the specified content appears in an image.
Davidsonian Scene Graph (DSG)~\citep{cho2024davidsonian} organizes atomic questions into dependency graphs to improve semantic coverage and answer consistency.
VQAScore~\citep{lin2024evaluating} measures image--text alignment through the probability of a ``Yes'' answer to a question asking whether the image depicts the supplied text.
I-HallA~\citep{lim2025evaluating} evaluates factual correctness in text-to-image generation using VQA accuracy on curated multiple-choice questions informed by factual knowledge and visual evidence.
Building on this question-based evaluation paradigm, VHA compares answer log-probabilities across GT and SR outputs using same question-answer pairs derived independently from each image.



\section{VQA-based Hallucination Assessment}
\label{sec:vqa_assessment}
\begin{figure*}[t]
    \centering
    \includegraphics[width=\textwidth]{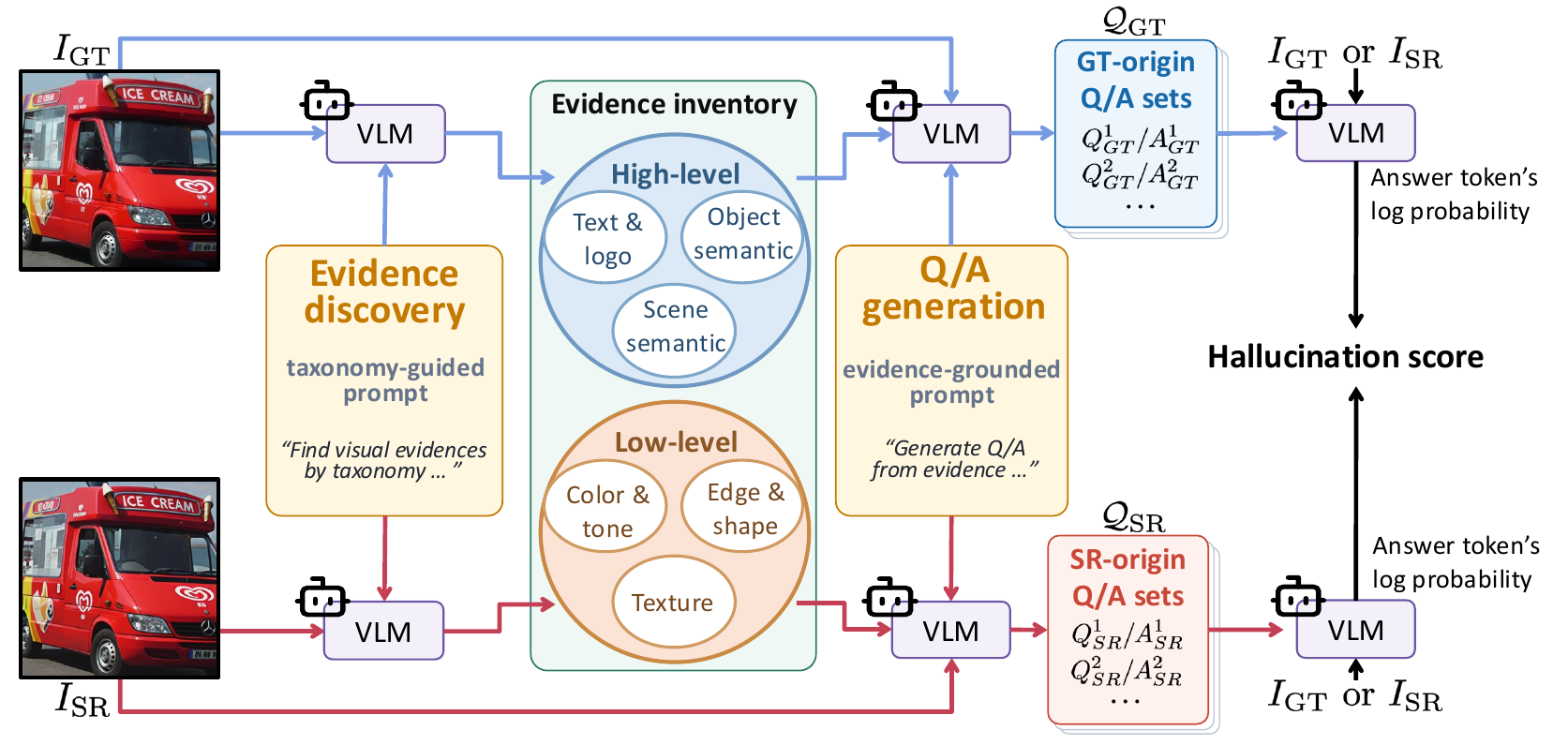}
    \caption{
        Overview of our bidirectional VQA-based hallucination assessment.
        Taxonomy-guided question--answer pairs are generated from both ground truth and super-resolved images, and each answer is scored under both images to assess evidence preservation and unsupported content.
    }
    \label{fig:overview_scoring}
\end{figure*}

We introduce VQA-based Hallucination Assessment (VHA) to quantify hallucinations in a super-resolved image ($I_{\mathrm{SR}}$) using its ground-truth counterpart ($I_{\mathrm{GT}}$) as the reference.
To examine diverse semantic content and visual attributes, we generate taxonomy-guided question–answer (Q/A) pairs from both images (Section~\ref{sec:qa_generation}) using a vision-language model (VLM).
For each Q/A pair, we compare the VLM’s answer log-probabilities under the two images on same Q/A pairs. 
GT-derived pairs assess the loss or alteration of original evidence, while SR-derived pairs assess unsupported additions; we aggregate these probability differences into a bidirectional hallucination score  (Section~\ref{sec:bidirectional_scoring}).
Figure~\ref{fig:overview_scoring} illustrates our assessment pipeline.

\subsection{Taxonomy-guided question--answer generation}
\label{sec:qa_generation}
We first generate Q/A pairs covering both high-level semantics and low-level visual attributes using a VLM.
Without explicit guidance on what to examine, a VLM may focus its questions on certain types of content and attributes while overlooking others.
We therefore use a taxonomy to guide Q/A generation across different types of visual evidence.
The high-level categories comprise scene semantics, object semantics, and text or logos, while the low-level categories comprise color and tone, edges and shapes, and texture.

We organize Q/A generation around explicitly identified visual targets to limit repeated questions about the same content.
Within each category, we first prompt the VLM to identify visible content or attributes.
We then prompt it to generate one question and a short answer for each identified item.
We apply this procedure independently to $I_{\mathrm{GT}}$ and $I_{\mathrm{SR}}$ to obtain the corresponding Q/A sets, $\mathcal{Q}_{\mathrm{GT}}$ and $\mathcal{Q}_{\mathrm{SR}}$, respectively.
The full prompts for both stages are provided in Appendix~\ref{app:qa_prompts}.

\subsection{Bidirectional hallucination scoring}
\label{sec:bidirectional_scoring}

We assess hallucinations by evaluating whether GT content is preserved in SR and whether SR content is supported by GT. 
Forward and reverse scoring use $\mathcal{Q}_{\mathrm{GT}}$ and $\mathcal{Q}_{\mathrm{SR}}$, respectively, to compare each same answer’s likelihood under both images.
For a question ($q$) and an answer consisting of $T$ tokens ($a=(a_1,\ldots,a_T)$), we compute the mean token-level log-probability under a fixed VLM ($p_\phi$) in a teacher-forced manner:
\begin{equation}
\ell(I;q,a)=\frac{1}{T}\sum_{t=1}^{T}
\log p_\phi(a_t\mid I,q,a_{<t}).
\label{eq:answer_logprob}
\end{equation}
%

We define the answer score as the difference between the image-conditioned and text-only answer log-probabilities:
\begin{equation}
s(I;q,a)=\ell(I;q,a)-\ell(\varnothing;q,a),
\label{eq:answer_support}
\end{equation}
where $\varnothing$ denotes an input without an image. 
We then use each Q/A pair’s score on its source image to gate its contribution to the hallucination score.

To assess the preservation of original ground-truth content and attributes, we define the forward score $F$ by aggregating score differences over GT-derived Q/A pairs:
\begin{equation}
F(I_{\mathrm{GT}},I_{\mathrm{SR}})=\frac{1}{|\mathcal{Q}_{\mathrm{GT}}|}
\sum_{(q,a)\in\mathcal{Q}_{\mathrm{GT}}}
\mathbf{1}\!\left[s(I_{\mathrm{GT}};q,a)>\tau_g\right]
\cdot \bigl[s(I_{\mathrm{GT}};q,a)-s(I_{\mathrm{SR}};q,a)\bigr].
\label{eq:forward_score}
\end{equation}
Here, $\mathbf 1[\cdot]$ is the indicator function, and $\tau_g$ is the GT answer-score threshold. 
For pairs passing the threshold, a decrease in the answer score from GT to SR contributes positively to $F$. 
These contributions can offset one another.

GT-derived questions may overlook newly introduced content in the SR output. 
We therefore define the reverse score $R$ to assess potentially unsupported additions by measuring decreases in answer scores from SR to GT for SR-derived Q/A pairs:
%
\begin{equation}
\begin{split}
R(I_{\mathrm{GT}},I_{\mathrm{SR}})=\frac{1}{|\mathcal{Q}_{\mathrm{SR}}|}
\sum_{(q,a)\in\mathcal{Q}_{\mathrm{SR}}}
&\mathbf{1}\!\left[s(I_{\mathrm{SR}};q,a)>\tau_s\right] \cdot \bigl[s(I_{\mathrm{SR}};q,a)-s(I_{\mathrm{GT}};q,a)-\tau_m\bigr]_+,
\end{split}
\label{eq:reverse_score}
\end{equation}
where $[x]_+=\max(x,0)$.
The SR answer-score threshold $\tau_s$ suppresses contributions from answers weakly supported by the SR output. 
The reverse score then counts only the portion of the SR–GT score difference exceeding the margin $\tau_m$.

The final hallucination score is computed as follows:
\begin{equation}
H(I_{\mathrm{GT}},I_{\mathrm{SR}})=F(I_{\mathrm{GT}},I_{\mathrm{SR}})+R(I_{\mathrm{GT}},I_{\mathrm{SR}}).
\label{eq:hallucination_score}
\end{equation}

\section{Learning a Lightweight Hallucination Assessment Model}
\label{sec:vha_training}
\begin{figure*}[t]
    \centering
    \scalebox{0.95}{
        \includegraphics[width=\textwidth]{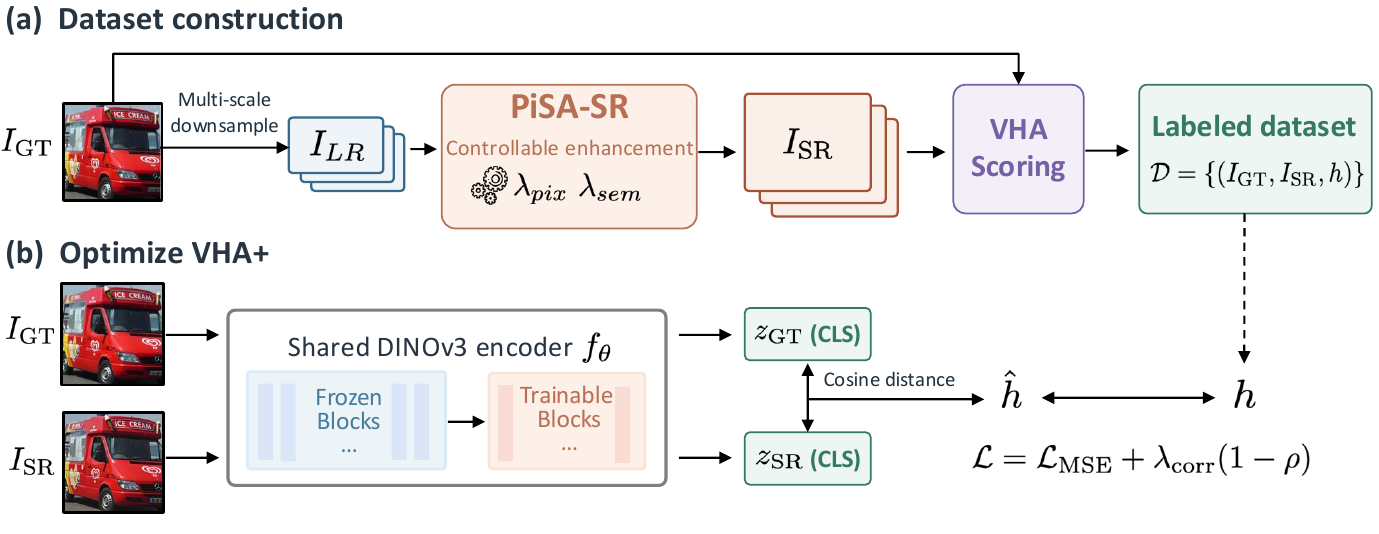}
    }
    \caption{
        Overview of our VHA+ training pipeline.
        (a) Diverse GT--SR pairs are constructed using PiSA-SR and labeled with our VHA scores.
        (b) VHA+ is trained to predict these scores directly from image pairs.
    }
    \label{fig:overview_vha}
\end{figure*}

VHA requires repeated VLM inference, and its full pipeline is not end-to-end differentiable. 
We therefore train VHA+, a lightweight, differentiable model that predicts VHA scores directly from GT–SR image pairs. 
Figure~\ref{fig:overview_vha} illustrates dataset construction (Section~\ref{sec:vha_dataset}) and model training (Section~\ref{sec:vha_prediction_training}).

\subsection{Training dataset construction}
\label{sec:vha_dataset}
We construct GT-SR pairs spanning different degrees of content preservation and alteration by varying the input resolution and enhancement settings.
Each GT image is downsampled at multiple scales and restored with PiSA-SR~\citep{sun2025pixel}, whose pixel-level and semantic-level controls vary fidelity and perceptual detail.
We label each pair with its VHA score to obtain the training dataset $\mathcal{D} = \{(I_{\mathrm{GT}}^{(i)}, I_{\mathrm{SR}}^{(i)}, h_i)\}_{i=1}^{N}$, where $h_i = H(I_{\mathrm{GT}}^{(i)}, I_{\mathrm{SR}}^{(i)})$ and $N$ is the number of image pairs.
Additional dataset construction details are provided in Section~\ref{sec:setup_detail}.



\subsection{Model architecture and training objective}
\label{sec:vha_prediction_training}
VHA+ predicts the VHA score from the distance between learned representations of the GT image and SR outputs.
Following the DINO-based scoring approach of Hallucination Score~\citep{ren2025hallucination}, we use a shared DINOv3 encoder $f_\theta$~\citep{simeoni2025dinov3} to extract the final CLS tokens, $z_{\mathrm{GT}} = f_{\theta}(I_{\mathrm{GT}})$ and $z_{\mathrm{SR}} = f_{\theta}(I_{\mathrm{SR}})$.
An affine transformation maps their cosine distance to the VHA score scale:
\begin{equation}
\hat{h}(I_{\mathrm{GT}},I_{\mathrm{SR}})=\alpha\left(1-
\frac{{z_{\mathrm{GT}}}^{\mathsf{T}}z_{\mathrm{SR}}}{\|z_{\mathrm{GT}}\|_2\|z_{\mathrm{SR}}\|_2}\right)+\beta,
\label{eq:vha_prediction}
\end{equation}
where $\alpha$ and $\beta$ are learned scale and offset parameters.

We fine-tune the final transformer blocks and learn the affine parameters, while keeping the earlier blocks frozen.
Supervision from VHA encourages the learned feature distance to reflect content loss, alteration, and unsupported additions.
The training objective combines mean squared error (MSE) on standardized scores with a Pearson correlation loss:
\begin{equation}
\mathcal{L}=\mathcal{L}_{\mathrm{MSE}}(\hat{\mathbf{h}},\mathbf{h})
+\lambda_{\mathrm{corr}}\bigl(1-\rho(\hat{\mathbf{h}},\mathbf{h})\bigr),
\label{eq:vha_loss}
\end{equation}
where $\hat{\mathbf{h}}$ and $\mathbf{h}$ contain the predicted and target VHA scores, $\rho$ is the Pearson correlation, and $\lambda_{\mathrm{corr}}$ controls the correlation term.
The MSE term encourages accurate prediction of target values, while the correlation term encourages the alignment in score distribution.
Additional training details are provided in Appendix~\ref{app:vha_training}.

\section{Experiments}
\label{sec:exp}

\subsection{Experimental Setup}
\label{sec:setup_detail}

\paragraph{Benchmark construction.}
We construct VHA-bench, a human-annotated hallucination benchmark, by collecting human judgments of hallucination in generative SR outputs.
For this study, we use 90 ground-truth images from the RealSR validation set~\citep{cai2019toward}.
Each $512\times512$ GT image is downsampled to $128\times128$, and the low-resolution input is restored to $512\times512$ using six generative SR methods: StableSR~\citep{wang2024exploiting}, DiffBIR~\citep{lin2024diffbir}, SeeSR~\citep{wu2024seesr}, OSEDiff~\citep{wu2024one}, PURE~\citep{wei2025perceive} and VOSR~\citep{wu2026vosr}, yielding 540 GT--SR pairs.
We conduct a human study with 200 participants, each randomly assigned 54 GT--SR pairs for evaluation.
Using the GT image as reference, participants rate hallucination on a five-point scale, from 1 (no hallucination) to 5 (severe hallucination), focusing on changes in semantic content and visual attributes while ignoring differences in sharpness or noise alone.
Each GT--SR pair receives approximately 20 ratings on average, which we aggregate into a mean opinion score (MOS).
The study interface is shown in Figure~\ref{fig:web_ui} in the appendix.

\begin{figure*}[t]
    \centering
    \begingroup

    \noindent\begin{minipage}[c]{0.33\textwidth}
        \centering
        \setlength{\tabcolsep}{1pt}
        \renewcommand{\arraystretch}{1.05}
        \scalebox{0.82}{%
            \begin{tabular}{@{}cc@{}}
                $I_{\mathrm{GT}}$                                                                                                  & $I_{\mathrm{SR}}$ (OSEDiff) \\
                \includegraphics[width=0.58\linewidth]{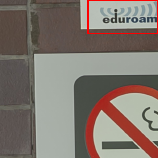} &
                \includegraphics[width=0.58\linewidth]{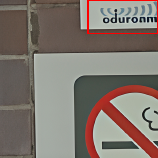}                               \\[2pt]
                \includegraphics[width=0.58\linewidth]{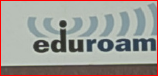}                       &
                \includegraphics[width=0.58\linewidth]{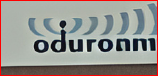}
            \end{tabular}%
        }
    \end{minipage}\hfill
    \begin{minipage}[c]{0.65\textwidth}
        \centering
        \setlength{\tabcolsep}{2pt}
        \renewcommand{\arraystretch}{1.05}
        \scalebox{0.75}{%
            \begin{tabular}{@{}lp{0.82\linewidth}rr@{}}
                \toprule
                \begin{tabular}[c]{@{}l@{}}Taxonomy /\\Source\end{tabular}\cellcolor{black!8}                         & \cellcolor{black!8} Question / Answer                                                         & \cellcolor{black!8} $s(I_{\mathrm{GT}})$ & \cellcolor{black!8} $s(I_{\mathrm{SR}})$ \\
                \midrule
                \begin{tabular}[t]{@{}l@{}}Text (high)\\$\mathcal{Q}_{\mathrm{GT}}$\end{tabular}   & \textbf{Q.} What text is on the upper sign?\newline\textbf{A.} eduroam    & $10.58$             & $-0.04$             \\
                \addlinespace[3pt]
                \begin{tabular}[t]{@{}l@{}}Text (high)\\$\mathcal{Q}_{\mathrm{SR}}$\end{tabular}   & \textbf{Q.} What text is on the upper sign?\newline\textbf{A.} oduronni   & $-2.15$             & $19.75$             \\
                \addlinespace[3pt]
                \begin{tabular}[t]{@{}l@{}}Object (high)\\$\mathcal{Q}_{\mathrm{GT}}$\end{tabular} & \textbf{Q.} What is the lower sign?\newline\textbf{A.} no smoking sign    & $5.04$              & $6.66$              \\
                \addlinespace[3pt]
                \begin{tabular}[t]{@{}l@{}}Scene (high)\\$\mathcal{Q}_{\mathrm{GT}}$\end{tabular}  & \textbf{Q.} What is the background surface?\newline\textbf{A.} brick wall & $9.12$              & $9.25$              \\
                \bottomrule
            \end{tabular}%
        }
    \end{minipage}
    \par\vspace{2pt}
    \noindent\begin{minipage}[c]{0.33\textwidth}
        \centering
        \setlength{\tabcolsep}{1pt}
        \renewcommand{\arraystretch}{1.05}
        \scalebox{0.82}{%
            \begin{tabular}{@{}cc@{}}
                $I_{\mathrm{GT}}$                                                                                                  & $I_{\mathrm{SR}}$ (OSEDiff) \\
                \includegraphics[width=0.58\linewidth]{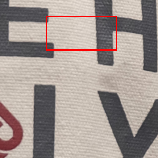} &
                \includegraphics[width=0.58\linewidth]{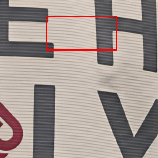}                               \\[2pt]
                \includegraphics[width=0.58\linewidth]{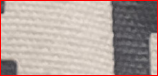}                    &
                \includegraphics[width=0.58\linewidth]{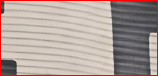}
            \end{tabular}%
        }
    \end{minipage}\hfill
    \begin{minipage}[c]{0.65\textwidth}
        \centering
        \setlength{\tabcolsep}{2pt}
        \renewcommand{\arraystretch}{1.05}
        \scalebox{0.75}{%
            \begin{tabular}{@{}lp{0.82\linewidth}rr@{}}
                \toprule
                \begin{tabular}[c]{@{}l@{}}Taxonomy /\\Source\end{tabular}\cellcolor{black!8}                         & \cellcolor{black!8} Question / Answer                                                                                 & \cellcolor{black!8} $s(I_{\mathrm{GT}})$ & \cellcolor{black!8} $s(I_{\mathrm{SR}})$ \\
                \midrule
                \begin{tabular}[t]{@{}l@{}}Text (high)\\$\mathcal{Q}_{\mathrm{GT}}$\end{tabular}   & \textbf{Q.} What letter is visible in upper left quadrant?\newline\textbf{A.} E                   & $13.00$             & $13.00$             \\
                \addlinespace[3pt]
                \begin{tabular}[t]{@{}l@{}}Object (high)\\$\mathcal{Q}_{\mathrm{GT}}$\end{tabular} & \textbf{Q.} What letter is in upper right?\newline\textbf{A.} H                                   & $11.90$             & $11.90$             \\
                \addlinespace[3pt]
                \begin{tabular}[t]{@{}l@{}}Texture (low)\\$\mathcal{Q}_{\mathrm{GT}}$\end{tabular} & \textbf{Q.} What does the black ink surface look like?\newline\textbf{A.} Slightly grainy texture & $1.97$              & $0.64$              \\
                \addlinespace[3pt]
                \begin{tabular}[t]{@{}l@{}}Texture (low)\\$\mathcal{Q}_{\mathrm{SR}}$\end{tabular} & \textbf{Q.} What is the background texture?\newline\textbf{A.} Fine horizontal ribs               & $-3.69$             & $2.90$              \\
                \bottomrule
            \end{tabular}%
        }
    \end{minipage}
    \par\vspace{2pt}
    \noindent\begin{minipage}[c]{0.33\textwidth}
        \centering
        \setlength{\tabcolsep}{1pt}
        \renewcommand{\arraystretch}{1.05}
        \scalebox{0.82}{%
            \begin{tabular}{@{}cc@{}}
                $I_{\mathrm{GT}}$                                                                                                  & $I_{\mathrm{SR}}$ (DiffBIR) \\
                \includegraphics[width=0.58\linewidth]{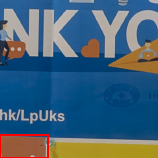} &
                \includegraphics[width=0.58\linewidth]{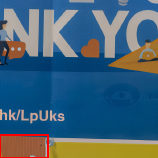}                               \\[2pt]
                \includegraphics[width=0.58\linewidth]{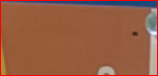}                       &
                \includegraphics[width=0.58\linewidth]{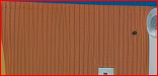}
            \end{tabular}%
        }
    \end{minipage}\hfill
    \begin{minipage}[c]{0.65\textwidth}
        \centering
        \setlength{\tabcolsep}{2pt}
        \renewcommand{\arraystretch}{1.05}
        \scalebox{0.75}{%
            \begin{tabular}{@{}lp{0.82\linewidth}rr@{}}
                \toprule
                \begin{tabular}[c]{@{}l@{}}Taxonomy /\\Source\end{tabular}\cellcolor{black!8}                         & \cellcolor{black!8} Question / Answer                                                                        & \cellcolor{black!8} $s(I_{\mathrm{GT}})$ & \cellcolor{black!8} $s(I_{\mathrm{SR}})$ \\
                \midrule
                \begin{tabular}[t]{@{}l@{}}Object (high)\\$\mathcal{Q}_{\mathrm{GT}}$\end{tabular} & \textbf{Q.} What is on left side of banner?\newline\textbf{A.} Walking person            & $3.72$              & $3.66$              \\
                \addlinespace[3pt]
                \begin{tabular}[t]{@{}l@{}}Object (high)\\$\mathcal{Q}_{\mathrm{GT}}$\end{tabular} & \textbf{Q.} What is on right side of banner?\newline\textbf{A.} Person on paper airplane & $6.59$              & $4.33$              \\
                \addlinespace[3pt]
                \begin{tabular}[t]{@{}l@{}}Tone (low)\\$\mathcal{Q}_{\mathrm{GT}}$\end{tabular} & \textbf{Q.} What contrast exists between text and background?\newline\textbf{A.} High contrast white on blue & $2.13$ & $1.90$ \\
                \addlinespace[3pt]
                \begin{tabular}[t]{@{}l@{}}Texture (low)\\$\mathcal{Q}_{\mathrm{SR}}$\end{tabular} & \textbf{Q.} What texture is on orange panel?\newline\textbf{A.} Vertical wood grain      & $-4.32$             & $2.85$              \\
                \bottomrule
            \end{tabular}%
        }
    \end{minipage}
    \endgroup
    \caption{
    Examples of Q/A generation and scoring on VHA-bench.
    $\mathcal{Q}_{\mathrm{GT}}$ and $\mathcal{Q}_{\mathrm{SR}}$ denote Q/A sets generated from the GT image and SR outputs, respectively.
    For each Q/A pair, $s(I)$ denotes the answer score $s(I;q,a)$ computed by Eq.~\ref{eq:answer_support}.
    }
    \label{fig:qa_qualitative}
\end{figure*}

\paragraph{Evaluation metrics.}
We use existing evaluation metrics to examine their correlations with human hallucination judgments on VHA-bench and to assess the outputs of SR methods.
The metrics include PSNR and SSIM for pixel-level similarity, and LPIPS~\citep{zhang2018unreasonable}, DISTS~\citep{ding2020image}, DINO, and CLIP for feature-level similarity.
For DINO and CLIP, we compute cosine distances between the final-layer CLS features of GT and SR images using frozen pretrained \texttt{DINOv3 ViT-B/16} and \texttt{CLIP ViT-B/16} encoders~\citep{simeoni2025dinov3, radford2021learning}, respectively.
For no-reference image quality assessment (NR-IQA), we include MUSIQ~\citep{ke2021musiq}, CLIP-IQA~\citep{wang2023exploring}, Q-Insight~\citep{li2026q}, VisualQuality-R1~\citep{wu2026visualquality}, and DeQA-Score~\citep{you2025teaching}.
We also include A-FINE~\citep{chen2025toward} as a generalized full-reference IQA (GFR-IQA) metric.

\paragraph{Implementation details.}
We use \texttt{Qwen3-VL-8B-Instruct}~\citep{bai2025qwen3} for Q/A generation and answer scoring.
For bidirectional VHA scoring, we set $\tau_g=-2$, $\tau_s=0$ and $\tau_m=0.5$.
For VHA+ training dataset construction, we select 20,000 images each from LSDIR~\citep{li2023lsdir} and TextVQA~\citep{singh2019towards} and use their $512\times512$ center crops as GT images.
We downsample each GT image by factors of $0.5$, $0.375$, and $0.25$ and restore the degraded inputs using PiSA-SR~\citep{sun2025pixel}, with and without additional $2\times$ input upsampling.
We set the pixel-level and semantic-level control weights of PiSA-SR $(\lambda_{\mathrm{pix}},\lambda_{\mathrm{sem}})$ to $(1,0)$, $(0.5,1)$, and $(1,1)$.

\subsection{Results}
\label{sec:exp_results}

\paragraph{Examples of Q/A generation and scoring.}
Figure~\ref{fig:qa_qualitative} illustrates how generated Q/A pairs capture high-level semantic content and low-level visual attributes, and how their score differences reflect changes between the GT image and SR outputs.
In the first example, the Q/A pair about the original text in $\mathcal{Q}_{\mathrm{GT}}$ yields a higher $s(I_{\mathrm{GT}})$ than $s(I_{\mathrm{SR}})$, whereas the pair about the hallucinated text in $\mathcal{Q}_{\mathrm{SR}}$ yields a lower $s(I_{\mathrm{GT}})$ than $s(I_{\mathrm{SR}})$.
In the second, the Q/A pair about texture in $\mathcal{Q}_{\mathrm{GT}}$ yields a higher $s(I_{\mathrm{GT}})$ than $s(I_{\mathrm{SR}})$, whereas the pair about background texture in $\mathcal{Q}_{\mathrm{SR}}$ yields a lower $s(I_{\mathrm{GT}})$ than $s(I_{\mathrm{SR}})$.
The Q/A pairs about unchanged letters receive nearly identical scores across the two images.
In the third, a Q/A pair in $\mathcal{Q}_{\mathrm{SR}}$ captures newly introduced wood-grain texture, with $s(I_{\mathrm{GT}})$ much lower than $s(I_{\mathrm{SR}})$.

\paragraph{Comparison of correlations with existing metrics.}
    
Table~\ref{tab:realsr_4x_human_correlation_twocol} shows that VHA and VHA+ achieve higher correlations with human hallucination judgments than the existing metrics evaluated on VHA-bench. 
VHA achieves Spearman and Pearson correlations of 0.618 and 0.645, respectively, compared with at most 0.282 and 0.299 among the existing metrics.
The low correlations of these baselines suggest that general image similarity and quality measures alone are insufficient to capture human judgments of hallucination severity. 
VHA+ achieves correlations of 0.659 and 0.675, respectively, similar to those of VHA. 
These results demonstrate that the lightweight predictor retains strong agreement with human judgments without requiring VLM inference at evaluation time.


\begin{table}[t]
    \centering
    \caption{Spearman ($\rho$) and Pearson ($r$) correlations with human mean opinion scores (MOS) for hallucination, computed on 540 SR outputs from six methods and rated by 200 participants. Higher correlations indicate closer agreement with human judgments. VHA denotes our bidirectional VQA-based assessment method, while VHA+ is its learned predictor that operates without VLM inference.}
    \label{tab:realsr_4x_human_correlation_twocol}
    \vspace{-2mm}
    \begingroup
    \small
    \setlength{\tabcolsep}{3pt}
    \setlength{\aboverulesep}{0pt}
    \setlength{\belowrulesep}{0pt}
    \renewcommand{\arraystretch}{1.16}
    \scalebox{.9}{
        \begin{tabular}{@{}p{0.105\textwidth}p{0.145\textwidth}
            >{\centering\arraybackslash}p{0.12\textwidth}
            >{\centering\arraybackslash}p{0.11\textwidth}
            @{\hspace{0.03\textwidth}}p{0.13\textwidth}p{0.15\textwidth}
            >{\centering\arraybackslash}p{0.11\textwidth}
            >{\centering\arraybackslash}p{0.10\textwidth}@{}}
            \toprule
            \addlinespace[2pt]
            \rowcolor{black!8}
            Category                        & Method                                               & \shortstack{$\rho\,\uparrow$} & \shortstack{$r\,\uparrow$}
                                            & Category                                             & Method                        & \shortstack{$\rho\,\uparrow$} & \shortstack{$r\,\uparrow$}                                \\
            \cmidrule(r{0.03\textwidth}){1-4}\cmidrule(l{0pt}){5-8}
            \addlinespace[2pt]
            \multirow{2}{*}{\footnotesize\textit{Pixel sim.}}
                                            & PSNR                                                 & 0.162                         & 0.181
                                            & \multirow{4}{*}{\footnotesize\textit{Feature sim.}}
                                            & LPIPS                                                & 0.083                         & 0.089                                                                                     \\
                                            & SSIM                                                 & 0.010                         & 0.012                         &                            & DISTS        & 0.066 & 0.069 \\
            \cmidrule(r{0.03\textwidth}){1-4}
            \multirow{5}{*}{\footnotesize\textit{NR-IQA}}
                                            & MUSIQ                                                & -0.008                        & -0.014                        &                            & DINO         & 0.081 & 0.115 \\
                                            & CLIP-IQA                                             & 0.168                         & 0.159                         &                            & CLIP         & 0.282 & 0.273 \\
            \cmidrule[\lightrulewidth](l{0pt}){5-8}
                                            & Q-Insight                                            & 0.080                         & 0.036
                                            & \multirow[c]{4}{0.13\textwidth}{\raggedright\footnotesize\textit{Hallucination}\\\textit{Assessment}}
                                            & Direct rating                                        & 0.497                         & 0.482                                                                                     \\
                                            & VQ-R1                                                & 0.175                         & 0.126                         &                            & HS prompting & 0.543 & 0.563 \\
                                            & DeQA                                                 & 0.145                         & 0.156
                                            &
                                            & \cellcolor{blue!5}\textbf{VHA (Ours)}
                                            & \cellcolor{blue!5}\underline{0.618}
                                            & \cellcolor{blue!5}\underline{0.645}                                                                                                                                              \\
            \cmidrule(r{0.03\textwidth}){1-4}
            {\footnotesize\textit{GFR-IQA}} & A-FINE                                               & 0.173                         & 0.299                         & 
                                            & \cellcolor{blue!5}\textbf{VHA+ (Ours)}
                                            & \cellcolor{blue!5}\textbf{0.659}
                                            & \cellcolor{blue!5}\textbf{0.675}                                                                                                                                                 \\
            \addlinespace[2pt]
            \bottomrule
        \end{tabular}
    }

    \endgroup
\end{table}

\paragraph{Comparison with  hallucination assessment methods.}
Table~\ref{tab:realsr_4x_human_correlation_twocol} compares hallucination assessment methods using the same VLM backbone (\texttt{Qwen3-VL-8B-Instruct}).
Direct rating provides the VLM with GT images and SR outputs and prompts it to assign a hallucination rating from 1 to 5.
The full prompt is provided in Appendix~\ref{app:direct_rating_prompts}.
HS prompting uses exactly the same prompt as Hallucination Score~\citep{ren2025hallucination}, with GT images, low-resolution inputs, and SR outputs.
Both baselines show reasonable correlations with human judgments, but their discrete ratings provide limited granularity. 
Also, neither baseline explicitly quantifies how visual elements or attributes contribute to the hallucination score.
Our method achieves higher correlations while providing continuous scores from answer log-probabilities and explicit contributions from individual Q/A pairs.

\subsection{Ablation studies and analysis}

\begin{table}[t]
    \centering
    \caption{
        Ablations of VHA scoring on VHA-bench.
        $\rho$ and $r$ denote Spearman and Pearson correlations.
    }
    \vspace{-2.5mm}
    \label{tab:realsr_4x_scoring_ablation}
    \captionsetup[subtable]{skip=3pt,justification=centering}
    \begin{subtable}[c]{0.3\textwidth}
        \centering
        \caption{Bidirectional scoring}
        \label{tab:realsr_4x_bidirectional_ablation}

        \setlength{\tabcolsep}{7pt}
        \renewcommand{\arraystretch}{1.15}
        \scalebox{.82}{%
        \begin{tabular}{lcc}
            \hline
            \rowcolor{black!8}
            Scoring & $\rho\,\uparrow$ & $r\,\uparrow$ \\
            \hline
            Forward & 0.516 & 0.562 \\
            Reverse & 0.465 & 0.560 \\
            Bidirectional & \textbf{0.618} & \textbf{0.645} \\
            \hline
        \end{tabular}%
        }
    \end{subtable}\hfill
    \begin{subtable}[c]{0.33\textwidth}
        \centering
        \caption{High- and low-level Q/A}
        \label{tab:realsr_4x_granularity_human}

        \setlength{\tabcolsep}{7pt}
        \renewcommand{\arraystretch}{1.15}
        \scalebox{.82}{%
        \begin{tabular}{lcc}
            \hline
            \rowcolor{black!8}
            Q/A set & $\rho\,\uparrow$ & $r\,\uparrow$ \\
            \hline
            High only & 0.539 & 0.584 \\
            Low only & 0.204 & 0.197 \\
            High + Low & \textbf{0.618} & \textbf{0.645} \\
            \hline
        \end{tabular}%
        }
    \end{subtable}\hfill
    \begin{subtable}[c]{0.33\textwidth}
        \centering
        \caption{Forward and reverse thresholding}
        \label{tab:realsr_4x_filtering_ablation}

        \setlength{\tabcolsep}{5.5pt}
        \renewcommand{\arraystretch}{1.15}
        \scalebox{.82}{%
        \begin{tabular}{cc|cc}
            \hline
            \rowcolor{black!8}
            Forward & Reverse & $\rho\,\uparrow$ & $r\,\uparrow$ \\
            \hline
            $\checkmark$ & & 0.570 & 0.614 \\
             & $\checkmark$ & 0.607 & 0.635 \\
            $\checkmark$ & $\checkmark$ & \textbf{0.618} & \textbf{0.645} \\
            \hline
        \end{tabular}%
        }
    \end{subtable}
\end{table}

\paragraph{Ablations of VHA scoring.}
Table~\ref{tab:realsr_4x_scoring_ablation} presents ablation results for VHA scoring on VHA-bench.
Table~\ref{tab:realsr_4x_scoring_ablation}(\subref{tab:realsr_4x_bidirectional_ablation}) shows that bidirectional scoring achieves higher correlations with human judgments than either forward or reverse scoring alone.
Table~\ref{tab:realsr_4x_scoring_ablation}(\subref{tab:realsr_4x_granularity_human}) shows that scores from high-level Q/A pairs align more closely with human hallucination judgments than those from low-level pairs.
Combining both further improves correlations, suggesting that high-level semantic content and low-level visual attributes provide complementary information for assessing hallucination.
Finally, Table~\ref{tab:realsr_4x_scoring_ablation}(\subref{tab:realsr_4x_filtering_ablation}) evaluates forward thresholding with $\tau_g$ in Eq.~\ref{eq:forward_score} and reverse thresholding with $\tau_s$ and $\tau_m$ in Eq.~\ref{eq:reverse_score}.
Both help suppress unreliable Q/A contributions and improve correlations with human judgments.

\begin{figure*}[t]
    \centering
    \includegraphics[width=\textwidth]{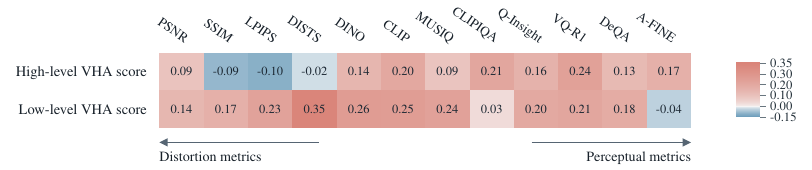}
    \vspace{-6mm}
    \caption{
        Spearman correlations of our high-level and low-level bidirectional Q/A scores with the image evaluation metrics on VHA-bench.
    }
    \label{fig:realsr_4x_granularity_heatmap}
\end{figure*}

\paragraph{Correlations of high- and low-level VHA scores with existing metrics.}
We compute high- and low-level VHA scores using only high- and low-level Q/A pairs, respectively.
Figure~\ref{fig:realsr_4x_granularity_heatmap} shows their Spearman correlations with existing metrics on VHA-bench.
The high-level score correlates weakly with pixel-level similarity metrics (PSNR and SSIM) and perceptual similarity metrics (LPIPS and DISTS), while showing relatively higher correlations with semantic feature distances (DINO and CLIP) and most NR-IQA metrics.
Compared with the high-level score, the low-level score shows higher correlations with most existing metrics, especially for pixel-level and perceptual similarity metrics.
This pattern suggests that many existing metrics reflect low-level visual changes more closely than high-level semantic changes.

\begin{wraptable}{r}{0.4\textwidth}
    \vspace{-13pt}
    \centering
    \caption{Parameter counts and average inference time of VHA and VHA+.}
    \vspace{-2mm}
    \label{tab:vha_runtime}
    \setlength{\tabcolsep}{3.5pt}
    \renewcommand{\arraystretch}{1.15}
    \scalebox{0.85}{%
        \begin{tabular}{lcc}
            \toprule
            \rowcolor{black!8}
            Method & Parameters (M) & Inference time (ms) \\
            \midrule
            VHA    & 8,767          & 101,000             \\
            VHA+   & 85             & 17                  \\
            \bottomrule
                    \vspace{-7mm}
        \end{tabular}%
    }
\end{wraptable}

\paragraph{Computational efficiency of VHA+.}
Table~\ref{tab:vha_runtime} compares the parameter counts and inference times of VHA and VHA+.
We measure the average inference time per image pair over 50 pairs on a single NVIDIA RTX A6000.
VHA+ uses 85M parameters, compared with 8,767M for VHA, and reduces the average inference time from 101\,s to 17\,ms per pair.
This corresponds to approximately $5{,}900\times$ faster inference under the measured configuration.
By directly predicting scores without VLM inference, VHA+ substantially reduces evaluation cost while retaining strong agreement with human judgments.

\subsection{Fine-tuning Super-Resolution with VHA+}
\label{sec:sr_finetuning}
The lightweight and differentiable design of VHA+ allows it to guide generative SR training.
We use the hallucination score predicted by the frozen VHA+ model as an auxiliary loss.
Following Hallucination Score~\citep{ren2025hallucination}, we combine this loss with a MUSIQ-based quality objective~\citep{ke2021musiq} to discourage oversmoothing and preserve perceptual quality.
We fine-tune pretrained VOSR~\citep{wu2026vosr} by adding these terms to its original distillation objective while retaining its training pipeline.
We denote the fine-tuned model as VOSR$^*$.
Training details, including datasets, loss weights, and optimization settings, are provided in Appendix~\ref{app:sr_training_details}.

\begin{table}[t]
    \centering
    \caption{Quantitative comparison of SR methods on RealSR and DRealSR.
        VOSR$^*$ denotes VOSR fine-tuned with auxiliary losses based on our VHA+ score and MUSIQ.}
    \label{tab:vosr_finetuning}
    \vspace{-2mm}
    \setlength{\tabcolsep}{2pt}
    \renewcommand{\arraystretch}{1.25}
    \aboverulesep=0pt
    \belowrulesep=0pt
    \resizebox{\linewidth}{!}{%
        \begin{tabular}{ll|*{8}{>{\centering\arraybackslash}p{0.12\linewidth}}|*{2}{>{\centering\arraybackslash}p{0.12\linewidth}}}
            \hline
            \rowcolor{black!8}
                                      &                          & \multicolumn{2}{c}{\textit{Pixel sim.}} & \multicolumn{2}{c}{\textit{Feature sim.}} & \multicolumn{3}{c}{\textit{NR-IQA}} & \multicolumn{1}{c|}{\textit{GFR-IQA}} & \multicolumn{2}{c}{\textit{Ours}}                                                                                                                        \\
            \cmidrule(lr){3-4}\cmidrule(lr){5-6}\cmidrule(lr){7-9}\cmidrule(lr){10-10}\cmidrule(lr){11-12}
            \rowcolor{black!8}
            \multirow{-2}{*}{Dataset} & \multirow{-2}{*}{Method} & PSNR $\uparrow$                         & SSIM $\uparrow$                           & LPIPS $\downarrow$                  & DISTS $\downarrow$                    & MUSIQ $\uparrow$                  & CLIPIQA $\uparrow$ & VQ-R1 $\uparrow$ & A-FINE $\downarrow$ & \textbf{VHA} $\downarrow$ & \textbf{VHA+} $\downarrow$ \\
            \hline
            \multirow{8}{*}{RealSR}   & Bicubic                  & 25.48                                   & 0.7200                                    & 0.4564                              & 0.2633                                & 25.81                             & 0.3098             & 1.2210           & 77.22               & 0.9861                    & 0.8055                     \\
                                      & StableSR                 & 23.08                                   & 0.6611                                    & 0.3107                              & 0.2186                                & 65.00                             & 0.6142             & 3.5362           & 27.48               & 1.4394                    & 0.9863                     \\
                                      & DiffBIR                  & 23.27                                   & 0.6140                                    & 0.3617                              & 0.2419                                & 68.30                             & 0.6921             & 3.9381           & 31.87               & 1.3234                    & 1.0500                     \\
                                      & SeeSR                    & 23.86                                   & 0.6879                                    & 0.2980                              & 0.2222                                & 69.49                             & 0.6725             & 3.8458           & 25.84               & 1.3439                    & 1.0230                     \\
                                      & OSEDiff                  & 23.59                                   & 0.6923                                    & 0.2921                              & 0.2128                                & 69.08                             & 0.6688             & 4.0439           & 20.96               & 1.2806                    & 1.0086                     \\
                                      & PURE                     & 21.52                                   & 0.5755                                    & 0.3829                              & 0.2450                                & 67.05                             & 0.6943             & 3.7221           & 41.68               & 1.3613                    & 1.1788                     \\
            \noalign{\vskip 0.2pt}
            \noalign{\global\arrayrulewidth=1pt}
            \cdashline{2-12}[2pt/1.5pt]
            \noalign{\global\arrayrulewidth=0.7pt}
            \noalign{\vskip 0.2pt}
                                      & VOSR                     & 23.77                                   & 0.6816                                    & 0.2858                              & 0.2097                                & 70.00                             & 0.5761             & 3.7426           & 21.39               & 1.0190                    & 0.9283                     \\
            \rowcolor{red!6}
            \cellcolor{white}         & VOSR$^*$ (Ours)          & \textbf{24.29}                          & \textbf{0.6914}                           & \textbf{0.2820}                     & \textbf{0.1985}                       & \textbf{70.98}                    & \textbf{0.6074}    & \textbf{3.8557}  & \textbf{11.24}      & \textbf{0.9494}           & \textbf{0.6413}            \\
            \hline
            \multirow{8}{*}{DRealSR}  & Bicubic                  & 27.87                                   & 0.7869                                    & 0.4608                              & 0.2788                                & 22.59                             & 0.3190             & 1.0280           & 74.54               & 0.7310                    & 0.6948                     \\
                                      & StableSR                 & 25.78                                   & 0.6985                                    & 0.3379                              & 0.2295                                & 58.32                             & 0.6250             & 3.1017           & 37.97               & 0.8453                    & 0.7859                     \\
                                      & DiffBIR                  & 23.52                                   & 0.5466                                    & 0.4923                              & 0.2913                                & 65.99                             & 0.7213             & 3.7510           & 33.10               & 1.2837                    & 1.1691                     \\
                                      & SeeSR                    & 26.09                                   & 0.7284                                    & 0.3205                              & 0.2304                                & 64.86                             & 0.6841             & 3.5276           & 22.73               & 1.1364                    & 0.9278                     \\
                                      & OSEDiff                  & 25.85                                   & 0.7374                                    & 0.2967                              & 0.2163                                & 64.69                             & 0.6961             & 3.7727           & 21.38               & 0.8258                    & 0.8162                     \\
                                      & PURE                     & 23.34                                   & 0.5762                                    & 0.4526                              & 0.2719                                & 61.08                             & 0.6773             & 3.3473           & 41.07               & 1.0318                    & 1.0200                     \\
            \noalign{\vskip 0.2pt}
            \noalign{\global\arrayrulewidth=1pt}
            \cdashline{2-12}[2pt/1.5pt]
            \noalign{\global\arrayrulewidth=0.7pt}
            \noalign{\vskip 0.2pt}
                                      & VOSR                     & 25.44                                   & 0.7029                                    & 0.3479                              & 0.2362                                & 65.80                             & 0.5906             & 3.4163           & 29.46               & 0.7516                    & 0.9781                     \\
            \rowcolor{red!6}
            \cellcolor{white}         & VOSR$^*$ (Ours)          & \textbf{25.83}                          & \textbf{0.7076}                           & \textbf{0.3473}                     & \textbf{0.2263}                       & \textbf{66.97}                    & \textbf{0.6422}    & \textbf{3.5343}  & \textbf{21.63}      & \textbf{0.7161}           & \textbf{0.6473}            \\
            \hline
        \end{tabular}%
    }
\end{table}

Table~\ref{tab:vosr_finetuning} compares VOSR$^*$ with pretrained generative SR models and bicubic upsampling on the RealSR~\citep{cai2019toward} and DRealSR~\citep{wei2020component} test sets provided by StableSR~\citep{wang2024exploiting}.
Although bicubic upsampling achieves high pixel-level similarity score, it produces blurry outputs which results in low perceptual quality score.
Also, it does not synthesize new fine details, thereby limiting hallucinations arising from unsupported additions.
Consistently, bicubic upsampling achieves better VHA and VHA+ scores than the pretrained generative SR models.
VOSR$^*$ improves all reported metrics over VOSR on both datasets, including VHA and VHA+.
These results suggest that combining VHA+-based supervision with a perceptual quality objective can reduce hallucination while maintaining perceptual quality.

\begin{figure}[t]
    \centering
    \begin{minipage}[t]{0.108\linewidth}
        \centering
        \includegraphics[width=\linewidth]{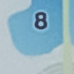}\par\vspace{1pt}
        \scriptsize GT
    \end{minipage}\hfill%
    \begin{minipage}[t]{0.108\linewidth}
        \centering
        \includegraphics[width=\linewidth]{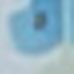}\par\vspace{1pt}
        \scriptsize Bicubic
    \end{minipage}\hfill%
    \begin{minipage}[t]{0.108\linewidth}
        \centering
        \includegraphics[width=\linewidth]{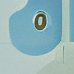}\par\vspace{1pt}
        \scriptsize StableSR
    \end{minipage}\hfill%
    \begin{minipage}[t]{0.108\linewidth}
        \centering
        \includegraphics[width=\linewidth]{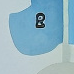}\par\vspace{1pt}
        \scriptsize DiffBIR
    \end{minipage}\hfill%
    \begin{minipage}[t]{0.108\linewidth}
        \centering
        \includegraphics[width=\linewidth]{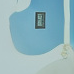}\par\vspace{1pt}
        \scriptsize SeeSR
    \end{minipage}\hfill%
    \begin{minipage}[t]{0.108\linewidth}
        \centering
        \includegraphics[width=\linewidth]{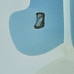}\par\vspace{1pt}
        \scriptsize OSEDiff
    \end{minipage}\hfill%
    \begin{minipage}[t]{0.108\linewidth}
        \centering
        \includegraphics[width=\linewidth]{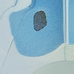}\par\vspace{1pt}
        \scriptsize PURE
    \end{minipage}\hfill%
    \begin{minipage}[t]{0.108\linewidth}
        \centering
        \includegraphics[width=\linewidth]{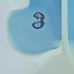}\par\vspace{1pt}
        \scriptsize VOSR
    \end{minipage}\hfill%
    \begin{minipage}[t]{0.108\linewidth}
        \centering
        \includegraphics[width=\linewidth]{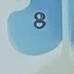}\par\vspace{1pt}
        \scriptsize VOSR$^*$(Ours)
    \end{minipage}%
    \par\vspace{5pt}
    \begin{minipage}[t]{0.108\linewidth}
        \centering
        \includegraphics[width=\linewidth]{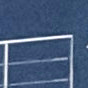}\par\vspace{1pt}
        \scriptsize GT
    \end{minipage}\hfill%
    \begin{minipage}[t]{0.108\linewidth}
        \centering
        \includegraphics[width=\linewidth]{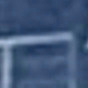}\par\vspace{1pt}
        \scriptsize Bicubic
    \end{minipage}\hfill%
    \begin{minipage}[t]{0.108\linewidth}
        \centering
        \includegraphics[width=\linewidth]{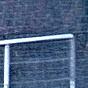}\par\vspace{1pt}
        \scriptsize StableSR
    \end{minipage}\hfill%
    \begin{minipage}[t]{0.108\linewidth}
        \centering
        \includegraphics[width=\linewidth]{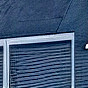}\par\vspace{1pt}
        \scriptsize DiffBIR
    \end{minipage}\hfill%
    \begin{minipage}[t]{0.108\linewidth}
        \centering
        \includegraphics[width=\linewidth]{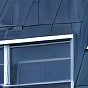}\par\vspace{1pt}
        \scriptsize SeeSR
    \end{minipage}\hfill%
    \begin{minipage}[t]{0.108\linewidth}
        \centering
        \includegraphics[width=\linewidth]{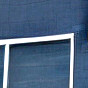}\par\vspace{1pt}
        \scriptsize OSEDiff
    \end{minipage}\hfill%
    \begin{minipage}[t]{0.108\linewidth}
        \centering
        \includegraphics[width=\linewidth]{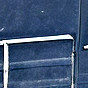}\par\vspace{1pt}
        \scriptsize PURE
    \end{minipage}\hfill%
    \begin{minipage}[t]{0.108\linewidth}
        \centering
        \includegraphics[width=\linewidth]{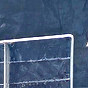}\par\vspace{1pt}
        \scriptsize VOSR
    \end{minipage}\hfill%
    \begin{minipage}[t]{0.108\linewidth}
        \centering
        \includegraphics[width=\linewidth]{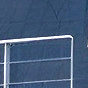}\par\vspace{1pt}
        \scriptsize VOSR$^*$(Ours)
    \end{minipage}%
    \caption{Qualitative comparisons of bicubic upsampling and generative SR methods on RealSR testset.
        VOSR$^*$, fine-tuned using our VHA+ score and MUSIQ as auxiliary losses, shows fewer hallucinated details and more faithfully preserves ground-truth content than other methods.}
    \label{fig:sr_qual_main}
    \vspace{-2mm}
\end{figure}

{Figure~\ref{fig:sr_qual_main} illustrates examples in which VOSR$^*$ better preserves text and suppresses spurious textures compared with VOSR.
In the top example, VOSR$^*$ more faithfully reconstructs the digit “8”, whose shape is distorted in the VOSR output. 
In the bottom example, VOSR$^*$ reduces spurious textures, more closely matching the smoother appearance of the GT image.}
Additional qualitative examples are provided in Appendix~\ref{app:sr_qual}.


\section{Conclusion}
\label{sec:conclusion}

We propose a bidirectional VQA-based method (VHA) for evaluating hallucination in generative image super-resolution.
Taxonomy-guided questions cover high-level semantic content and low-level visual attributes, enabling assessment of diverse forms of hallucination.
Bidirectional comparisons of answer score assess both the loss or alteration of ground-truth evidence and unsupported additions in SR outputs.
This addresses a gap in conventional SR evaluation, where visually realistic details can receive favorable quality scores despite misrepresenting the original content.
We also train VHA+, a lightweight model that directly predicts hallucination score without VLM inference. 
On our human-annotated benchmark, VHA and VHA+ achieve high correlations with human hallucination judgments.
Furthermore, our VHA+-based supervision extends this assessment to SR model training.
We believe this work adds a complementary dimension to SR evaluation and encourages future research to consider hallucination alongside perceptual quality when developing SR models.

\bibliography{iclr2027_conference}
\bibliographystyle{iclr2027_conference}

\newpage

\appendix

\section{Prompts for Q/A generation}
\label{app:qa_prompts}
Tables~\ref{tab:prompt_high_evidence}--\ref{tab:prompt_low_qa} provide the full prompt templates used for taxonomy-guided evidence discovery and evidence-conditioned question--answer generation in Section~\ref{sec:vqa_assessment}.
We use separate templates for high-level semantics and low-level visual attributes, and apply the same templates independently to ground truth and super-resolved images.

\begin{table}[!htbp]
\centering
\caption{High-level taxonomy-guided evidence discovery prompt.}
\label{tab:prompt_high_evidence}
\renewcommand{\arraystretch}{1.0}
\setlength{\tabcolsep}{0pt}
\scalebox{0.75}{%
\begin{tabular}{l}
\toprule
\begin{minipage}{1.25\linewidth}
\vspace{0.4em}
\ttfamily\raggedright
Look at this HR image (\{image\_id\}) and identify high-level semantic\par
evidence targets for short-answer Q/A generation.\par
\vspace{0.45em}
Taxonomy definitions:\par
\vspace{0.45em}
- scene\_semantic: The overall scene, environment, or image-level context.\par
\hspace*{1.0em}This taxonomy must describe the image as a whole or a broad scene region,\par
\hspace*{1.0em}not a single object.\par
\vspace{0.45em}
- object\_semantic: A distinct object or structure that contributes\par
\hspace*{1.0em}meaningfully to the image content, such as a tree, grass, tower, wall,\par
\hspace*{1.0em}house, vehicle, bridge, or signboard; or a recognizable visual property\par
\hspace*{1.0em}or part within an object or local region, such as texture, style,\par
\hspace*{1.0em}material, decoration, or a visible object part.\par
\vspace{0.45em}
- text\_logo: Clearly readable text, words, numbers, signs, brands, or logos.\par
\vspace{0.45em}
Generate up to 1 scene\_semantic target, up to 4 object\_semantic targets,\par
and up to 4 text\_logo targets.\par
\vspace{0.45em}
A target is suitable when it is clearly visible in the HR image,\par
supports a specific short-answer question, and may be altered by\par
downsampling or super-resolution.\par
\vspace{0.45em}
Generate as many valid targets as are clearly supported by the image,\par
within these limits.\par
\vspace{0.45em}
Reply with valid JSON only:\par
\vspace{0.45em}
[\par
\hspace*{1.0em}\{\par
\hspace*{2.0em}"evidence\_id": "he01",\par
\hspace*{2.0em}"taxonomy": "scene\_semantic",\par
\hspace*{2.0em}"region\_scope": "whole\_scene or localized region",\par
\hspace*{2.0em}"region\_hint": "brief visible location",\par
\hspace*{2.0em}"target": "specific high-level semantic content",\par
\hspace*{2.0em}"hallucination\_risk": "how this content may be altered or invented"\par
\hspace*{1.0em}\}\par
]\par
\vspace{0.4em}
\end{minipage} \\
\bottomrule
\end{tabular}%
}
\end{table}

\begin{table}[!htbp]
\centering
\caption{High-level evidence-conditioned question--answer generation prompt.}
\label{tab:prompt_high_qa}
\renewcommand{\arraystretch}{1.0}
\setlength{\tabcolsep}{0pt}
\scalebox{0.75}{%
\begin{tabular}{l}
\toprule
\begin{minipage}{1.25\linewidth}
\vspace{0.4em}
\ttfamily\raggedright
Look at this HR image (\{image\_id\}) and generate one high-level\par
short-answer Q/A pair for each provided \{taxonomy\}\par
evidence item.\par
\vspace{0.45em}
Taxonomy definition:\par
\vspace{0.45em}
\{taxonomy\_definition\}\par
\vspace{0.45em}
The question must identify the relevant image region unambiguously and\par
ask only about the specified high-level semantic content.\par
\vspace{0.45em}
Use a precise semantic label or short noun phrase.\par
\vspace{0.45em}
Do not ask about overall image quality.\par
\vspace{0.45em}
Do not create yes/no questions.\par
\vspace{0.45em}
Each answer must be specific and 1-3 words only.\par
\vspace{0.45em}
Evidence items:\par
\vspace{0.45em}
\{evidence\_json\}\par
\vspace{0.45em}
Reply with valid JSON only:\par
\vspace{0.45em}
[\par
\hspace*{1.0em}\{"evidence\_id": "\{example\_evidence\_id\}", "q": "...", "a": "..."\}\par
]\par
\vspace{0.4em}
\end{minipage} \\
\bottomrule
\end{tabular}%
}
\end{table}

\begin{table}[!htbp]
\centering
\caption{Low-level taxonomy-guided evidence discovery prompt.}
\label{tab:prompt_low_evidence}
\renewcommand{\arraystretch}{1.0}
\setlength{\tabcolsep}{0pt}
\scalebox{0.75}{%
\begin{tabular}{l}
\toprule
\begin{minipage}{1.25\linewidth}
\vspace{0.4em}
\ttfamily\raggedright
Look at this HR image (\{image\_id\}) and identify low-level visual evidence\par
targets for short-answer Q/A generation.\par
\vspace{0.45em}
Taxonomy definitions:\par
\vspace{0.45em}
- color\_tone: Visible color, brightness, contrast, highlight, shadow, or\par
\hspace*{1.0em}tone relationship within an image region.\par
\vspace{0.45em}
- edge\_shape: Visible orientation, curvature, thickness, continuity, or\par
\hspace*{1.0em}shape of a boundary or small structural part.\par
\vspace{0.45em}
- texture: Visible surface grain, material appearance, fine pattern,\par
\hspace*{1.0em}line structure, or texture density.\par
\vspace{0.45em}
Generate 3 color\_tone targets, 3 edge\_shape targets, and 3 texture targets\par
for this image. A target is suitable when it is clearly visible in the HR\par
image, supports a specific short-answer question, and may be altered by\par
downsampling or super-resolution.\par
\vspace{0.45em}
Reply with valid JSON only:\par
[\par
\hspace*{1.0em}\{\par
\hspace*{2.0em}"evidence\_id": "le01",\par
\hspace*{2.0em}"taxonomy": "color\_tone",\par
\hspace*{2.0em}"region\_scope": "local",\par
\hspace*{2.0em}"region\_hint": "brief visible location",\par
\hspace*{2.0em}"target": "specific low-level visual property",\par
\hspace*{2.0em}"hallucination\_risk": "how this property may be altered or invented"\par
\hspace*{1.0em}\}\par
]\par
\vspace{0.4em}
\end{minipage} \\
\bottomrule
\end{tabular}%
}
\end{table}

\begin{table}[!htbp]
\centering
\caption{Low-level evidence-conditioned question--answer generation prompt.}
\label{tab:prompt_low_qa}
\renewcommand{\arraystretch}{1.0}
\setlength{\tabcolsep}{0pt}
\scalebox{0.75}{%
\begin{tabular}{l}
\toprule
\begin{minipage}{1.25\linewidth}
\vspace{0.4em}
\ttfamily\raggedright
Look at this HR image (\{image\_id\}) and generate one highly specific\par
low-level short-answer Q/A pair for each provided evidence item.\par
\vspace{0.45em}
The question must identify the relevant image region and ask only about\par
the specified low-level visual property.\par
\vspace{0.45em}
Describe the finest reliably visible appearance. Avoid broad categorical\par
answers when the image supports a more precise description.\par
\vspace{0.45em}
Use the following attribute structure:\par
\vspace{0.45em}
- color\_tone: describe the visible hue together with its lightness,\par
\hspace*{1.0em}saturation, or local tonal character.\par
\vspace{0.45em}
- edge\_shape: describe the visible boundary using its thickness,\par
\hspace*{1.0em}continuity, smoothness, geometry, or orientation.\par
\vspace{0.45em}
- texture: describe the visible texture using its scale, direction,\par
\hspace*{1.0em}regularity, density, or structural appearance.\par
\vspace{0.45em}
Combine two or three complementary visual attributes when they are\par
clearly supported by the image. Do not add an attribute merely to make\par
the answer longer or more specific.\par
\vspace{0.45em}
Use visually grounded everyday terms. Do not use uncommon or technical\par
terminology solely for greater specificity.\par
\vspace{0.45em}
Do not ask about overall image quality or sharpness preference.\par
Do not create yes/no questions.\par
\vspace{0.45em}
Each answer must be specific and 2-3 words long.\par
\vspace{0.45em}
Evidence items:\par
\{evidence\_json\}\par
\vspace{0.45em}
Reply with valid JSON only:\par
[\par
\hspace*{1.0em}\{"evidence\_id": "le01", "q": "...", "a": "..."\}\par
]\par
\vspace{0.4em}
\end{minipage} \\
\bottomrule
\end{tabular}%
}
\end{table}

\clearpage
\section{VHA+ training details}
\label{app:vha_training}
We train the last four transformer blocks of the \texttt{DINOv3 ViT-B/16} encoder and the affine scale and offset parameters, while keeping the earlier layers frozen.
We set $\lambda_{\mathrm{corr}}=3$ in Eq.~\ref{eq:vha_loss} and train VHA+ using AdamW~\citep{loshchilov2018decoupled} for 20,000 steps with a batch size of 32.
The learning rates are $10^{-6}$ for the encoder and $10^{-4}$ for the affine parameters.

\paragraph{Score standardization and calibration.}
Let $\mu_H$ and $\sigma_H$ denote the mean and standard deviation of the training-set target scores.
We standardize each target score $h_i$ as
\begin{equation}
    \tilde{h}_i=\frac{h_i-\mu_H}{\sigma_H}.
    \label{eq:vha_standardization}
\end{equation}
We train the affine head of VHA+ to predict the standardized target $\tilde{h}_i$.
Both loss terms in Eq.~\ref{eq:vha_loss} are computed between the model outputs and the standardized targets $\tilde{h}_i$.
For numerical stability, we add $10^{-8}$ to the denominator when computing the Pearson correlation.
At inference, we convert the affine head output to the original score scale using the stored training-set mean and standard deviation.
For our training dataset, $\mu_H=0.7545$ and $\sigma_H=1.0254$.

\section{Human study interface}
\begin{figure*}[h]
    \centering
    \includegraphics[width=\textwidth]{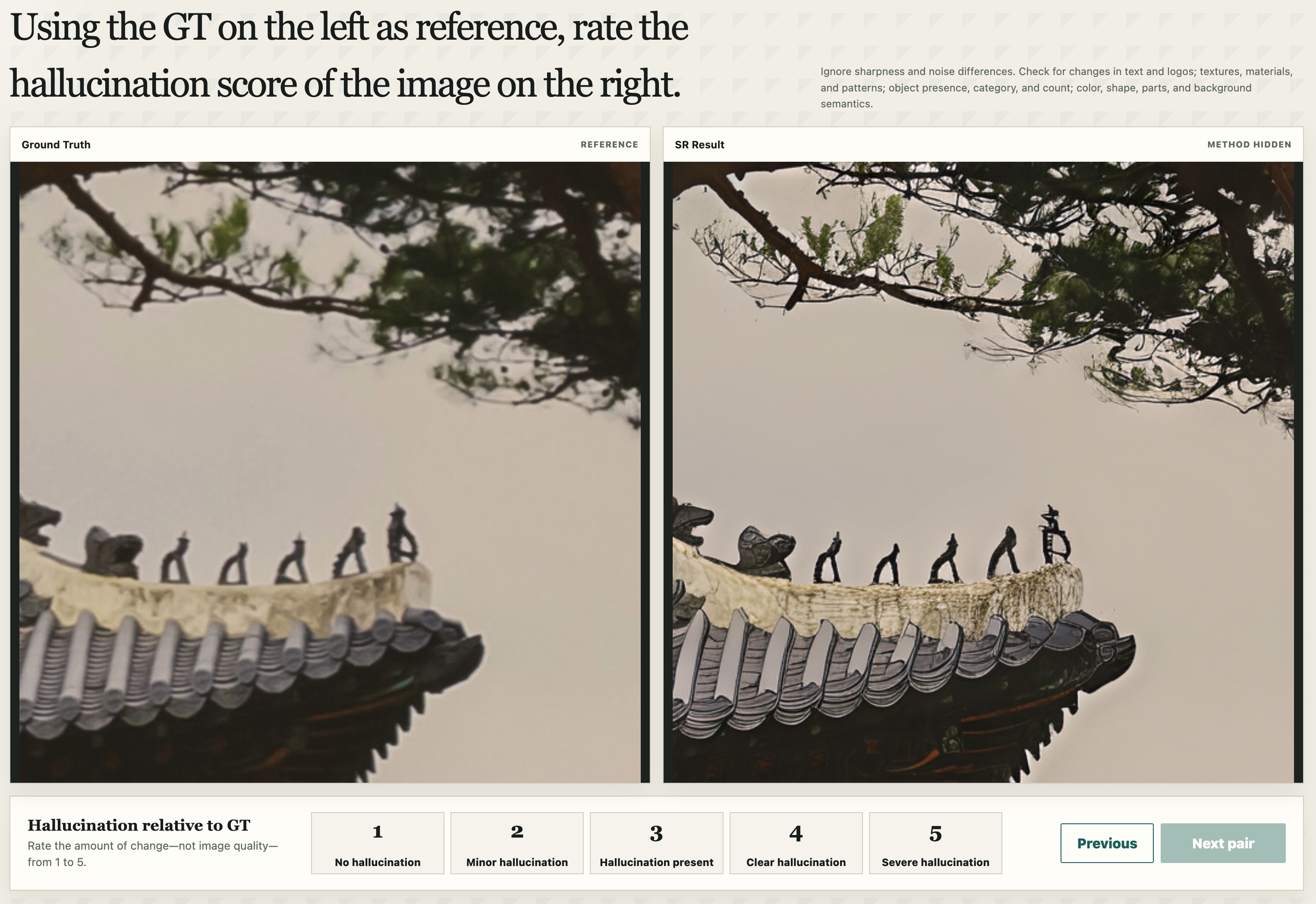}
    \caption{
        Web interface used for the human hallucination study.
        Participants compare the ground-truth reference (left) with a super-resolution result (right) and assign a hallucination score from 1 (no hallucination) to 5 (severe hallucination), while ignoring differences in sharpness and noise.
    }
    \label{fig:web_ui}
\end{figure*}


\clearpage
\section{Prompts for direct VLM rating}
\label{app:direct_rating_prompts}
Table~\ref{tab:prompt_direct_rating} provides the full prompt used for the Direct rating baseline in the comparison of VLM-based hallucination assessment methods in Section~\ref{sec:exp_results} (Table~\ref{tab:realsr_4x_human_correlation_twocol}).

\vspace{-2mm}
\begin{table}[!htbp]
    \centering
    \caption{Full prompt for Direct rating.}
    \vspace{-2mm}
    \label{tab:prompt_direct_rating}
    \renewcommand{\arraystretch}{1.0}
    \setlength{\tabcolsep}{0pt}
    \scalebox{0.75}{%
        \begin{tabular}{l}
            \toprule
            \begin{minipage}{1.25\linewidth}
                \vspace{0.4em}
                \ttfamily\raggedright
                You will receive two images:\par
                1. Ground Truth (GT): The reference high-resolution image.\par
                2. Super-Resolved Image (SR): An image restored by an AI super-resolution\par
                \hspace*{1.5em}model.\par
                \vspace{0.45em}
                We define hallucinations as the loss or alteration of ground-truth evidence in SR, or the addition of content unsupported by GT.\par
                \vspace{0.45em}
                Rate the amount of hallucination in SR compared with GT on a scale from 1 to 5.\par
                A lower score means less hallucination, and a higher score means more\par
                hallucination.\par
                \vspace{0.45em}
                Output only the following JSON:\par
                \{\par
                \hspace*{1.0em}"score": \textless{}integer from 1 to 5\textgreater{}\par
                \}\par
                \vspace{0.4em}
            \end{minipage} \\
            \bottomrule
        \end{tabular}%
    }
\end{table}
\vspace{-2mm}

\vspace{-2mm}
\section{Additional ablations}
\label{app:additional_ablations}
\subsection{VHA scoring threshold ablations}
\label{app:scoring_thresholds}

\vspace{-2mm}
\begin{table}[h]
    \centering
    \caption{Ablations of VHA scoring thresholds on VHA-bench.
    $\rho$ and $r$ denote Spearman and Pearson correlations.}
    \vspace{-2mm}
    \label{tab:realsr_4x_threshold_sweeps}
    \captionsetup[subtable]{skip=4pt,justification=centering}
    \setlength{\tabcolsep}{13pt}
    \renewcommand{\arraystretch}{1.0}
    \begin{subtable}[t]{0.33\textwidth}
        \centering
        \caption{Forward score threshold}
        \label{tab:forward_support_sweep}
        \scalebox{0.8}{%
        \begin{tabular}{rcc}
            \toprule
            \rowcolor{black!8}
            $\tau_g$ & $\rho$ & $r$ \\
            \midrule
            $0$ & 0.5919 & 0.6295 \\
            $-1$ & \textbf{0.6190} & 0.6427 \\
            $\mathbf{-2}$ & {0.6179} & \textbf{0.6449} \\
            $-3$ & 0.6103 & 0.6408 \\
            $-4$ & 0.6085 & 0.6396 \\
            $-5$ & 0.6075 & 0.6355 \\
            $-6$ & 0.6079 & 0.6354 \\
            \bottomrule
        \end{tabular}%
        }
    \end{subtable}\hfill
    \begin{subtable}[t]{0.33\textwidth}
        \centering
        \caption{Reverse score threshold}
        \label{tab:reverse_support_sweep}
        \scalebox{0.8}{%
        \begin{tabular}{rcc}
            \toprule
            \rowcolor{black!8}
            $\tau_s$ & $\rho$ & $r$ \\
            \midrule
            $-3$ & 0.6160 & 0.6426 \\
            $-2$ & 0.6164 & 0.6431 \\
            $-1$ & 0.6159 & 0.6437 \\
            $\mathbf{0}$ & \textbf{0.6179} & \textbf{0.6449} \\
            $1$ & 0.6144 & 0.6430 \\
            $2$ & 0.6087 & 0.6383 \\
            $3$ & 0.5941 & 0.6295 \\
            \bottomrule
        \end{tabular}%
        }
    \end{subtable}\hfill
    \begin{subtable}[t]{0.33\textwidth}
        \centering
        \caption{Reverse margin threshold}
        \label{tab:reverse_margin_sweep}
        \scalebox{0.8}{%
        \begin{tabular}{rcc}
            \toprule
            \rowcolor{black!8}
            $\tau_m$ & $\rho$ & $r$ \\
            \midrule
            $-1$ & 0.5909 & 0.6366 \\
            $-0.5$ & 0.6045 & 0.6407 \\
            $0$ & 0.6173 & 0.6443 \\
            $\mathbf{0.5}$ & \textbf{0.6179} & \textbf{0.6449} \\
            $1$ & 0.6133 & 0.6423 \\
            \bottomrule
        \end{tabular}%
        }
    \end{subtable}
\end{table}
\vspace{-2mm}

Table~\ref{tab:realsr_4x_threshold_sweeps} shows the effects of the GT answer-score threshold $\tau_g$ in Eq.~\ref{eq:forward_score} and the SR answer-score threshold $\tau_s$ and margin $\tau_m$ in Eq.~\ref{eq:reverse_score}.
VHA is relatively insensitive to variations in these hyperparameters.
However, overly strict or permissive thresholds can reduce correlations with human judgments.
These results suggest that appropriate thresholds help suppress noisy Q/A contributions while preserving informative ones.

\subsection{VHA+ training ablations}
\label{app:vha_training_ablation}

\begin{table}[h]
    \centering
    \caption{VHA+ training ablations on VHA-bench.
        $\rho$ and $r$ denote Spearman and Pearson correlations, respectively; bold indicates the selected settings.}
    \label{tab:vha_training_ablation}
    \vspace{-2mm}
    \captionsetup[subtable]{skip=3pt}
    \begin{subtable}[t]{0.48\textwidth}
        \centering
        \caption{Correlation loss weight}
        \label{tab:vha_pearson_weight}

        \setlength{\tabcolsep}{22pt}
        \renewcommand{\arraystretch}{1.15}
        \scalebox{.8}{%
        \begin{tabular}{ccc}
            \hline
            \rowcolor{black!8}
            $\lambda_{\mathrm{corr}}$ & $\rho$ & $r$ \\
            \hline
            0 & 0.6370 & 0.6680 \\
            1 & 0.6540 & 0.6771 \\
            2 & 0.6586 & 0.6763 \\
            \textbf{3} & \textbf{0.6594} & \textbf{0.6753} \\
            4 & 0.6593 & 0.6741 \\
            5 & 0.6582 & 0.6735 \\
            \hline
        \end{tabular}%
        }
    \end{subtable}\hfill
    \begin{subtable}[t]{0.48\textwidth}
        \centering
        \caption{Trainable blocks}
        \label{tab:vha_trainable_blocks}

        \setlength{\tabcolsep}{14pt}
        \renewcommand{\arraystretch}{1.15}
        \scalebox{.8}{%
        \begin{tabular}{ccc}
            \hline
            \rowcolor{black!8}
            Last $K$ blocks & $\rho$ & $r$ \\
            \hline
            2 & 0.5962 & 0.6212 \\
            \textbf{4} & \textbf{0.6594} & \textbf{0.6753} \\
            6 & 0.6556 & 0.6638 \\
            8 & 0.6494 & 0.6660 \\
            10 & 0.6389 & 0.6631 \\
            \hline
        \end{tabular}%
        }
    \end{subtable}
    \vspace{-2mm}
\end{table}

Table~\ref{tab:vha_training_ablation} presents VHA+ training ablations.
Table~\ref{tab:vha_training_ablation}(\subref{tab:vha_pearson_weight}) shows the results of varying the Pearson correlation loss weight $\lambda_{\mathrm{corr}}$ in Eq.~\ref{eq:vha_loss}.
An appropriate correlation loss weight improves agreement with human judgments compared with MSE alone.
Table~\ref{tab:vha_training_ablation}(\subref{tab:vha_trainable_blocks}) shows the results for different numbers of trainable transformer blocks.
Updating fewer or more than four blocks yields lower correlations, which may reflect underfitting and overfitting, respectively.

\section{Fine-Tuning SR Models with VHA+}
\label{app:sr_finetuning}

\subsection{Additional Training Details}
\label{app:sr_training_details}
VOSR~\citep{wu2026vosr} first trains a multi-step SR model and then distills it into a one-step student.
We fine-tune the pretrained one-step student with the additional VHA+ and MUSIQ losses while keeping the released multi-step model fixed as the teacher.
We use the full LSDIR training set~\citep{li2023lsdir} and all 800 DIV2K training images~\citep{agustsson2017ntire}.
Following VOSR's training setup, we use random $512\times512$ crops and its degradation pipeline.
The fine-tuning objective is
\begin{equation}
    \mathcal{L}
    =\mathcal{L}_{\mathrm{distill}}
    +\gamma_{\mathrm{vha}}\mathcal{L}_{\mathrm{VHA+}}
    +\gamma_{\mathrm{musiq}}\mathcal{L}_{\mathrm{MUSIQ}},
    \label{eq:sr_finetuning_loss}
\end{equation}
where $\mathcal{L}_{\mathrm{distill}}$ is VOSR's original distillation loss.
The auxiliary losses $\mathcal{L}_{\mathrm{VHA+}}$ and $\mathcal{L}_{\mathrm{MUSIQ}}$ are the score predicted by the frozen VHA+ model and the negative MUSIQ score~\citep{ke2021musiq}, respectively.
We set $\gamma_{\mathrm{vha}}=0.1$ and $\gamma_{\mathrm{musiq}}=0.001$.
We fine-tune the student for 1,000 updates using AdamW~\citep{loshchilov2018decoupled}, a learning rate of $10^{-5}$, and a batch size of 16.
We use an exponential moving average (EMA) with decay $0.99$.

\subsection{SR fine-tuning ablations}
\label{app:sr_loss_weights}
\begin{table}[t]
    \centering
    \caption{Ablations of VHA+ and MUSIQ loss weights for VOSR fine-tuning.}
    \label{tab:vosr_loss_weights}
    \setlength{\tabcolsep}{2pt}
    \renewcommand{\arraystretch}{1.25}
    \aboverulesep=0pt
    \belowrulesep=0pt
    \scalebox{0.64}{%
        \begin{tabular}{l|cc|*{8}{>{\centering\arraybackslash}p{0.12\linewidth}}|*{2}{>{\centering\arraybackslash}p{0.12\linewidth}}}
            \hline
            \rowcolor{black!8}
                                      &                                           &                                             & \multicolumn{2}{c}{\textit{Pixel sim.}} & \multicolumn{2}{c}{\textit{Feature sim.}} & \multicolumn{3}{c}{\textit{NR-IQA}} & \multicolumn{1}{c|}{\textit{GFR-IQA}} & \multicolumn{2}{c}{\textit{Ours}}                                                                                                      \\
            \cmidrule(lr){4-5}\cmidrule(lr){6-7}\cmidrule(lr){8-10}\cmidrule(lr){11-11}\cmidrule(lr){12-13}
            \rowcolor{black!8}
            \multirow{-2}{*}{Dataset} & \multirow{-2}{*}{$\gamma_{\mathrm{vha}}$} & \multirow{-2}{*}{$\gamma_{\mathrm{musiq}}$} & PSNR $\uparrow$                         & SSIM $\uparrow$                           & LPIPS $\downarrow$                  & DISTS $\downarrow$                    & MUSIQ $\uparrow$                  & CLIPIQA $\uparrow$ & VQ-R1 $\uparrow$ & A-FINE $\downarrow$ & VHA $\downarrow$ & VHA+ $\downarrow$ \\
            \hline
            \multirow{10}{*}{RealSR}  & 0                                         & 0.001                                       & 23.04                                   & 0.6418                                    & 0.3601                              & 0.2643                                & 75.89                             & 0.6820             & 3.723            & 11.29               & 1.372            & 1.172             \\
                                      & 0.05                                      & 0.001                                       & 23.96                                   & 0.6791                                    & 0.2925                              & 0.2089                                & 72.17                             & 0.6373             & 3.930            & 11.03               & 0.9164           & 0.6856            \\
            \cellcolor{white}         & 0.10                                      & 0.001                                       & 24.29                                   & 0.6914                                    & 0.2820                              & 0.1985                                & 70.98                             & 0.6074             & 3.856            & 11.24               & 0.9494           & 0.6413            \\
                                      & 0.15                                      & 0.001                                       & 24.25                                   & 0.6914                                    & 0.2698                              & 0.1907                                & 70.25                             & 0.5910             & 3.769            & 13.17               & 0.9557           & 0.6304            \\
                                      & 0.20                                      & 0.001                                       & 24.38                                   & 0.6988                                    & 0.2616                              & 0.1892                                & 69.39                             & 0.5610             & 3.791            & 13.94               & 0.9355           & 0.6211            \\
            \cdashline{2-13}[2pt/1.5pt]
                                      & 0.10                                      & 0                                           & 25.78                                   & 0.7431                                    & 0.2237                              & 0.1737                                & 51.31                             & 0.3457             & 2.709            & 48.03               & 0.8744           & 0.5645            \\
                                      & 0.10                                      & 0.0005                                      & 24.53                                   & 0.7026                                    & 0.2618                              & 0.1892                                & 68.87                             & 0.5710             & 3.761            & 16.08               & 0.9410           & 0.6341            \\
            \cellcolor{white}         & 0.10                                      & 0.0010                                      & 24.29                                   & 0.6914                                    & 0.2820                              & 0.1985                                & 70.98                             & 0.6074             & 3.856            & 11.24               & 0.9494           & 0.6413            \\
                                      & 0.10                                      & 0.0015                                      & 23.98                                   & 0.6816                                    & 0.2899                              & 0.2026                                & 71.83                             & 0.6255             & 3.943            & 10.10               & 1.0331           & 0.6536            \\
                                      & 0.10                                      & 0.0020                                      & 23.70                                   & 0.6653                                    & 0.3042                              & 0.2087                                & 73.02                             & 0.6391             & 3.945            & 8.811               & 0.9854           & 0.6641            \\
            \hline
            \multirow{10}{*}{DRealSR} & 0                                         & 0.0010                                      & 24.51                                   & 0.6482                                    & 0.4357                              & 0.3002                                & 73.57                             & 0.6899             & 3.184            & 12.27               & 1.271            & 1.256             \\
                                      & 0.05                                      & 0.0010                                      & 25.51                                   & 0.6954                                    & 0.3573                              & 0.2349                                & 68.52                             & 0.6545             & 3.678            & 18.59               & 0.7262           & 0.6788            \\
            \cellcolor{white}         & 0.10                                      & 0.0010                                      & 25.83                                   & 0.7076                                    & 0.3473                              & 0.2263                                & 66.97                             & 0.6422             & 3.534            & 21.63               & 0.7161           & 0.6473            \\
                                      & 0.15                                      & 0.0010                                      & 25.84                                   & 0.7060                                    & 0.3349                              & 0.2175                                & 66.01                             & 0.6237             & 3.471            & 25.25               & 0.6745           & 0.6101            \\
                                      & 0.20                                      & 0.0010                                      & 25.95                                   & 0.7139                                    & 0.3266                              & 0.2146                                & 65.11                             & 0.6042             & 3.459            & 27.07               & 0.6342           & 0.6043            \\
            \cdashline{2-13}[2pt/1.5pt]
                                      & 0.10                                      & 0                                           & 27.24                                   & 0.7714                                    & 0.2539                              & 0.1837                                & 48.03                             & 0.4260             & 2.489            & 52.57               & 0.6347           & 0.5812            \\
                                      & 0.10                                      & 0.0005                                      & 26.08                                   & 0.7173                                    & 0.3247                              & 0.2140                                & 64.51                             & 0.6123             & 3.462            & 28.58               & 0.7118           & 0.6473            \\
            \cellcolor{white}         & 0.10                                      & 0.0010                                      & 25.83                                   & 0.7076                                    & 0.3473                              & 0.2263                                & 66.97                             & 0.6422             & 3.534            & 21.63               & 0.7161           & 0.6473            \\
                                      & 0.10                                      & 0.0015                                      & 25.52                                   & 0.6947                                    & 0.3592                              & 0.2310                                & 68.30                             & 0.6541             & 3.666            & 20.93               & 0.6843           & 0.6469            \\
                                      & 0.10                                      & 0.0020                                      & 25.30                                   & 0.6799                                    & 0.3733                              & 0.2386                                & 69.49                             & 0.6648             & 3.673            & 19.24               & 0.6466           & 0.6527            \\
            \hline
        \end{tabular}%
    }
\end{table}

Table~\ref{tab:vosr_loss_weights} shows the results of varying the VHA+ and MUSIQ loss weights in Eq.~\ref{eq:sr_finetuning_loss}.
Increasing the VHA+ loss weight tends to lower perceptual-quality scores, whereas increasing the MUSIQ loss weight can reduce image similarity and increase VHA+ scores.
These results suggest that appropriate loss weights are needed to reduce hallucinations while maintaining perceptual quality.

\clearpage
\section{Additional qualitative results}
\label{app:sr_qual}
\begin{figure}[H]
    \centering
    \begin{minipage}[c]{0.3475\textwidth}
        \centering
        \includegraphics[width=\linewidth]{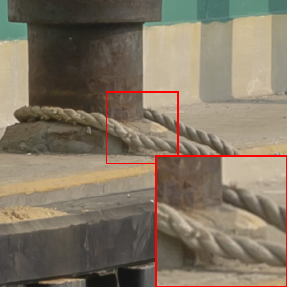}\par\vspace{-1pt}
        \small GT
    \end{minipage}\hfill%
    \begin{minipage}[c]{0.647\textwidth}
        \centering
        \begin{minipage}[t]{0.245\linewidth}
            \centering
            \includegraphics[width=\linewidth]{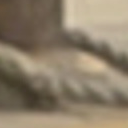}\par\vspace{-1pt}
            \small Bicubic
        \end{minipage}\hfill%
        \begin{minipage}[t]{0.245\linewidth}
            \centering
            \includegraphics[width=\linewidth]{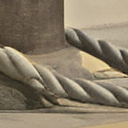}\par\vspace{-1pt}
            \small StableSR
        \end{minipage}\hfill%
        \begin{minipage}[t]{0.245\linewidth}
            \centering
            \includegraphics[width=\linewidth]{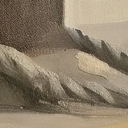}\par\vspace{-1pt}
            \small DiffBIR
        \end{minipage}\hfill%
        \begin{minipage}[t]{0.245\linewidth}
            \centering
            \includegraphics[width=\linewidth]{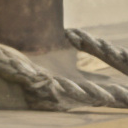}\par\vspace{-1pt}
            \small SeeSR
        \end{minipage}%
        \par\vspace{2pt}
        \begin{minipage}[t]{0.245\linewidth}
            \centering
            \includegraphics[width=\linewidth]{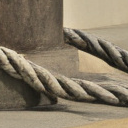}\par\vspace{-1pt}
            \small OSEDiff
        \end{minipage}\hfill%
        \begin{minipage}[t]{0.245\linewidth}
            \centering
            \includegraphics[width=\linewidth]{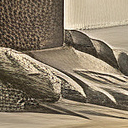}\par\vspace{-1pt}
            \small PURE
        \end{minipage}\hfill%
        \begin{minipage}[t]{0.245\linewidth}
            \centering
            \includegraphics[width=\linewidth]{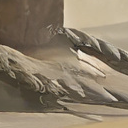}\par\vspace{-1pt}
            \small VOSR
        \end{minipage}\hfill%
        \begin{minipage}[t]{0.245\linewidth}
            \centering
            \includegraphics[width=\linewidth]{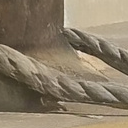}\par\vspace{-1pt}
            \small VOSR$^*$
        \end{minipage}%
    \end{minipage}
    \par\vspace{4pt}
    \begin{minipage}[c]{0.3475\textwidth}
        \centering
        \includegraphics[width=\linewidth]{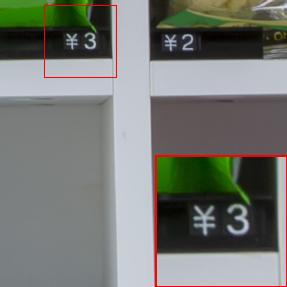}\par\vspace{-1pt}
        \small GT
    \end{minipage}\hfill%
    \begin{minipage}[c]{0.647\textwidth}
        \centering
        \begin{minipage}[t]{0.245\linewidth}
            \centering
            \includegraphics[width=\linewidth]{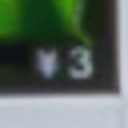}\par\vspace{-1pt}
            \small Bicubic
        \end{minipage}\hfill%
        \begin{minipage}[t]{0.245\linewidth}
            \centering
            \includegraphics[width=\linewidth]{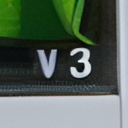}\par\vspace{-1pt}
            \small StableSR
        \end{minipage}\hfill%
        \begin{minipage}[t]{0.245\linewidth}
            \centering
            \includegraphics[width=\linewidth]{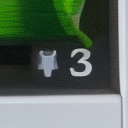}\par\vspace{-1pt}
            \small DiffBIR
        \end{minipage}\hfill%
        \begin{minipage}[t]{0.245\linewidth}
            \centering
            \includegraphics[width=\linewidth]{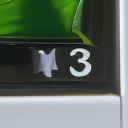}\par\vspace{-1pt}
            \small SeeSR
        \end{minipage}%
        \par\vspace{2pt}
        \begin{minipage}[t]{0.245\linewidth}
            \centering
            \includegraphics[width=\linewidth]{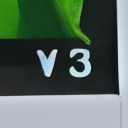}\par\vspace{-1pt}
            \small OSEDiff
        \end{minipage}\hfill%
        \begin{minipage}[t]{0.245\linewidth}
            \centering
            \includegraphics[width=\linewidth]{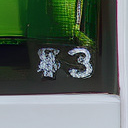}\par\vspace{-1pt}
            \small PURE
        \end{minipage}\hfill%
        \begin{minipage}[t]{0.245\linewidth}
            \centering
            \includegraphics[width=\linewidth]{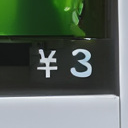}\par\vspace{-1pt}
            \small VOSR
        \end{minipage}\hfill%
        \begin{minipage}[t]{0.245\linewidth}
            \centering
            \includegraphics[width=\linewidth]{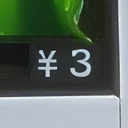}\par\vspace{-1pt}
            \small VOSR$^*$
        \end{minipage}%
    \end{minipage}
    \par\vspace{4pt}
    \begin{minipage}[c]{0.3475\textwidth}
        \centering
        \includegraphics[width=\linewidth]{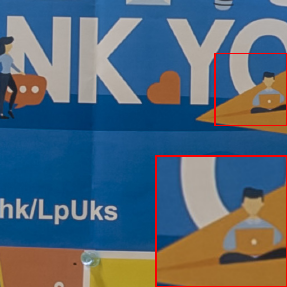}\par\vspace{-1pt}
        \small GT
    \end{minipage}\hfill%
    \begin{minipage}[c]{0.647\textwidth}
        \centering
        \begin{minipage}[t]{0.245\linewidth}
            \centering
            \includegraphics[width=\linewidth]{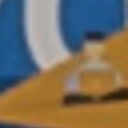}\par\vspace{-1pt}
            \small Bicubic
        \end{minipage}\hfill%
        \begin{minipage}[t]{0.245\linewidth}
            \centering
            \includegraphics[width=\linewidth]{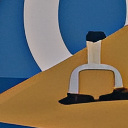}\par\vspace{-1pt}
            \small StableSR
        \end{minipage}\hfill%
        \begin{minipage}[t]{0.245\linewidth}
            \centering
            \includegraphics[width=\linewidth]{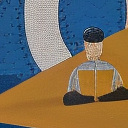}\par\vspace{-1pt}
            \small DiffBIR
        \end{minipage}\hfill%
        \begin{minipage}[t]{0.245\linewidth}
            \centering
            \includegraphics[width=\linewidth]{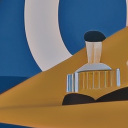}\par\vspace{-1pt}
            \small SeeSR
        \end{minipage}%
        \par\vspace{2pt}
        \begin{minipage}[t]{0.245\linewidth}
            \centering
            \includegraphics[width=\linewidth]{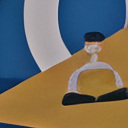}\par\vspace{-1pt}
            \small OSEDiff
        \end{minipage}\hfill%
        \begin{minipage}[t]{0.245\linewidth}
            \centering
            \includegraphics[width=\linewidth]{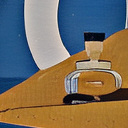}\par\vspace{-1pt}
            \small PURE
        \end{minipage}\hfill%
        \begin{minipage}[t]{0.245\linewidth}
            \centering
            \includegraphics[width=\linewidth]{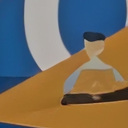}\par\vspace{-1pt}
            \small VOSR
        \end{minipage}\hfill%
        \begin{minipage}[t]{0.245\linewidth}
            \centering
            \includegraphics[width=\linewidth]{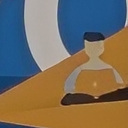}\par\vspace{-1pt}
            \small VOSR$^*$
        \end{minipage}%
    \end{minipage}
    \caption{Qualitative comparisons of bicubic upsampling and generative SR methods on RealSR testset.
        VOSR$^*$, fine-tuned using our VHA+ score and MUSIQ as auxiliary losses, shows fewer hallucinated details and more faithfully preserves ground-truth content than other methods.}
    \label{fig:sr_qual_Canon_043_LR4}
    \label{fig:sr_qual_Canon_047_LR4}
    \label{fig:sr_qual_Nikon_034_LR4}
\end{figure}

\begin{figure}[H]
    \centering
    \begin{minipage}[c]{0.3475\textwidth}
        \centering
        \includegraphics[width=\linewidth]{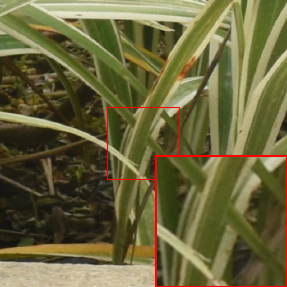}\par\vspace{-1pt}
        \small GT
    \end{minipage}\hfill%
    \begin{minipage}[c]{0.647\textwidth}
        \centering
        \begin{minipage}[t]{0.245\linewidth}
            \centering
            \includegraphics[width=\linewidth]{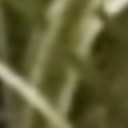}\par\vspace{-1pt}
            \small Bicubic
        \end{minipage}\hfill%
        \begin{minipage}[t]{0.245\linewidth}
            \centering
            \includegraphics[width=\linewidth]{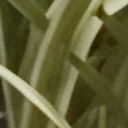}\par\vspace{-1pt}
            \small StableSR
        \end{minipage}\hfill%
        \begin{minipage}[t]{0.245\linewidth}
            \centering
            \includegraphics[width=\linewidth]{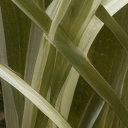}\par\vspace{-1pt}
            \small DiffBIR
        \end{minipage}\hfill%
        \begin{minipage}[t]{0.245\linewidth}
            \centering
            \includegraphics[width=\linewidth]{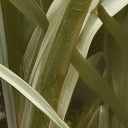}\par\vspace{-1pt}
            \small SeeSR
        \end{minipage}%
        \par\vspace{2pt}
        \begin{minipage}[t]{0.245\linewidth}
            \centering
            \includegraphics[width=\linewidth]{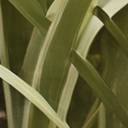}\par\vspace{-1pt}
            \small OSEDiff
        \end{minipage}\hfill%
        \begin{minipage}[t]{0.245\linewidth}
            \centering
            \includegraphics[width=\linewidth]{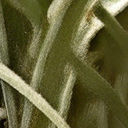}\par\vspace{-1pt}
            \small PURE
        \end{minipage}\hfill%
        \begin{minipage}[t]{0.245\linewidth}
            \centering
            \includegraphics[width=\linewidth]{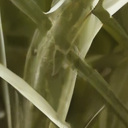}\par\vspace{-1pt}
            \small VOSR
        \end{minipage}\hfill%
        \begin{minipage}[t]{0.245\linewidth}
            \centering
            \includegraphics[width=\linewidth]{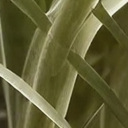}\par\vspace{-1pt}
            \small VOSR$^*$
        \end{minipage}%
    \end{minipage}
    \par\vspace{4pt}
    \begin{minipage}[c]{0.3475\textwidth}
        \centering
        \includegraphics[width=\linewidth]{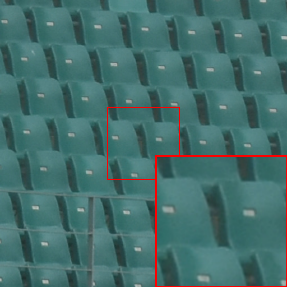}\par\vspace{-1pt}
        \small GT
    \end{minipage}\hfill%
    \begin{minipage}[c]{0.647\textwidth}
        \centering
        \begin{minipage}[t]{0.245\linewidth}
            \centering
            \includegraphics[width=\linewidth]{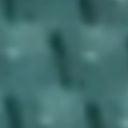}\par\vspace{-1pt}
            \small Bicubic
        \end{minipage}\hfill%
        \begin{minipage}[t]{0.245\linewidth}
            \centering
            \includegraphics[width=\linewidth]{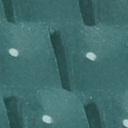}\par\vspace{-1pt}
            \small StableSR
        \end{minipage}\hfill%
        \begin{minipage}[t]{0.245\linewidth}
            \centering
            \includegraphics[width=\linewidth]{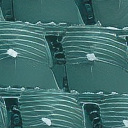}\par\vspace{-1pt}
            \small DiffBIR
        \end{minipage}\hfill%
        \begin{minipage}[t]{0.245\linewidth}
            \centering
            \includegraphics[width=\linewidth]{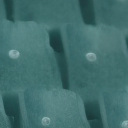}\par\vspace{-1pt}
            \small SeeSR
        \end{minipage}%
        \par\vspace{2pt}
        \begin{minipage}[t]{0.245\linewidth}
            \centering
            \includegraphics[width=\linewidth]{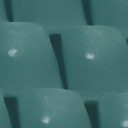}\par\vspace{-1pt}
            \small OSEDiff
        \end{minipage}\hfill%
        \begin{minipage}[t]{0.245\linewidth}
            \centering
            \includegraphics[width=\linewidth]{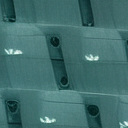}\par\vspace{-1pt}
            \small PURE
        \end{minipage}\hfill%
        \begin{minipage}[t]{0.245\linewidth}
            \centering
            \includegraphics[width=\linewidth]{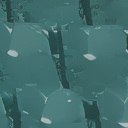}\par\vspace{-1pt}
            \small VOSR
        \end{minipage}\hfill%
        \begin{minipage}[t]{0.245\linewidth}
            \centering
            \includegraphics[width=\linewidth]{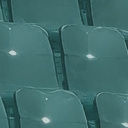}\par\vspace{-1pt}
            \small VOSR$^*$
        \end{minipage}%
    \end{minipage}
    \caption{Qualitative comparisons of bicubic upsampling and generative SR methods on DRealSR testset.
        VOSR$^*$, fine-tuned using our VHA+ score and MUSIQ as auxiliary losses, shows fewer hallucinated details and more faithfully preserves ground-truth content than pretrained VOSR.}
    \label{fig:sr_qual_DSC_1137_x1}
    \label{fig:sr_qual_DSC_1286_x1}
\end{figure}

Figures~\ref{fig:sr_qual_Canon_043_LR4} and~\ref{fig:sr_qual_DSC_1137_x1} present additional qualitative comparisons of generative SR models and bicubic upsampling on the RealSR~\citep{cai2019toward} and DRealSR~\citep{wei2020component} test sets provided by StableSR~\citep{wang2024exploiting}, respectively.

\clearpage

\end{document}